\documentclass[preprint,journal]{vgtc}       

\ifpdf
  \pdfoutput=1\relax                   
  \usepackage{graphicx}                
  \DeclareGraphicsExtensions{.pdf,.png,.jpg,.jpeg} 
\else
  \usepackage{graphicx}                
  \DeclareGraphicsExtensions{.eps}     
\fi%

\graphicspath{{figures/}{pictures/}{images/}{./}} 

\usepackage{microtype}                 
\PassOptionsToPackage{warn}{textcomp}  
\usepackage{textcomp}                  
\usepackage{mathptmx}                  
\usepackage{times}                     
\usepackage{cite}                      
\usepackage{tabu}                      
\usepackage{booktabs}                  

\usepackage{amsmath}
\usepackage{amssymb}
\usepackage{amsfonts}
\usepackage{subfig}
\usepackage{floatrow}
\usepackage{bm} 
\usepackage{helvet}
\usepackage[usenames, dvipsnames]{color}
\usepackage[font={scriptsize,sf}]{caption}
\usepackage[keeplastbox]{flushend}

\usepackage{multicol}
\preprinttext{}

\onlineid{1162}

\vgtccategory{Research}
\vgtcpapertype{algorithm/technique}

\title{Supporting Analysis of Dimensionality Reduction Results with Contrastive Learning
\vspace{-11pt}
}

\author{Takanori Fujiwara, Oh-Hyun Kwon, and Kwan-Liu Ma}
\authorfooter{
\item Takanori Fujiwara, Oh-Hyun Kwon, and Kwan-Liu Ma are with University of California, Davis. E-mail: \{tfujiwara, kw, klma\}@ucdavis.edu
}

\shortauthortitle{Fujiwara \MakeLowercase{\textit{et al.}}: \inserttitle}

\abstract{
Dimensionality reduction (DR) is frequently used for analyzing and visualizing high-dimensional data as it provides a good first glance of the data. 
However, to interpret the DR result for gaining useful insights from the data, it would take additional analysis effort such as identifying clusters and understanding their characteristics.
While there are many automatic methods (e.g., density-based clustering methods) to identify clusters, effective methods for understanding a cluster's characteristics are still lacking.
A cluster can be mostly characterized by its distribution of feature values. 
Reviewing the original feature values is not a straightforward task when the number of features is large. 
To address this challenge, we present a visual analytics method that effectively highlights the essential features of a cluster in a DR result.
To extract the essential features, we introduce an enhanced usage of contrastive principal component analysis (cPCA). 
Our method, called ccPCA (contrasting clusters in PCA), can calculate each feature's relative contribution to the \textit{contrast} between one cluster and other clusters. 
With ccPCA, we have created an interactive system including a scalable visualization of clusters' feature contributions. 
We demonstrate the effectiveness of our method and system with case studies using several publicly available datasets.
}


\CCScatlist{ 
    \CCScat{I.3.8}{Computer Graphics}{Applications}%
}

\vgtcinsertpkg

\definecolor{green}{rgb}{0,0,0}

\ieeedoi{10.1109/TVCG.2019.2934251}

\begin{document}

\firstsection{Introduction}
\maketitle

\captionsetup[algorithm]{font={scriptsize,sf}}
\setlength{\abovedisplayskip}{3pt}
\setlength{\belowdisplayskip}{3pt}

High-dimensional data visualization is one of the major research topics in the visualization community~\cite{liu2014survey,liu2017visualizing}.
Various types of visualization methods (e.g., the parallel coordinates~\cite{inselberg1987parallel}, scatterplot matrices~\cite{hartigan1975printer}, and star coordinates~\cite{kandogan2000star}) have been introduced to present high-dimensional information in a space~\cite{liu2017visualizing} (typically 2D on a computer screen) that human viewers can perceive and interpret.
Among these methods, dimensionality reduction (DR) methods are suitable to provide an overview of the relationships across the high-dimensional data points~\cite{liu2017visualizing,sacha2017visual,nonato2018multidimensional}.

The strength of DR methods is their capability of uncovering the similarity between data points as spatial proximity~\cite{wenskovitch2018towards}. 
In DR results, by referring to the ``similarity $\approx$ proximity''~\cite{wenskovitch2018towards} relationship, we can intuitively find useful patterns, such as clusters and outliers.
Many fields of study, including biology~\cite{hollt2018cyteguide}, social science~\cite{tsai2011dimensionality}, and machine learning~\cite{rauber2017visualizing}, require analyzing high-dimensional data and thus rely on DR methods.

According to the recent surveys~\cite{brehmer2014visualizing,nonato2018multidimensional}, analyzing a DR result involves the following tasks: (1) identifying clusters in the DR result, (2) understanding the characteristics of the clusters, and (3) comparing the clusters with predefined classes of data points~\cite{brehmer2014visualizing,nonato2018multidimensional}.
In the case that the DR result has interpretable axes, such as the dimensions generated by principal components analysis (PCA)~\cite{hotelling1933analysis,jolliffe1986principal}, understanding the characteristics of each axis and comparing the axis with the original dimensions (or features) are also included as part of the analysis tasks. 

Among the aforementioned tasks, the main task sequence is first identifying clusters and then understanding their characteristics~\cite{brehmer2014visualizing}.
While many automatic methods (e.g., density-based clustering methods~\cite{ester1996density,ankerst1999optics,kriegel2011density,campello2013density}) have been introduced to identify clusters (the first task), methods to assist the second task have still not been well studied, especially in the case that the data has many features. 
Reviewing the original feature values is essential to understanding each cluster's characteristics.
To support this task, many existing visual analytics systems~\cite{stahnke2016probing,pagliosa2016understanding,marcilio2017approach,demiralp2017clustrophile,kwon2018clustervision} employ basic statistical plots, such as histograms and parallel coordinates, for inspecting each feature of the selected clusters. 
However, because these visualizations render all of the features' values, they are limited in handling a large number of features. 
In addition, even if we were able to show all the features, it could be very time-consuming to find the common patterns within each cluster or find the differences among the clusters by individually referring to the values for each of the many associated features.

To address these problems, we have developed an analysis method that highlights those essential features for understanding characteristics of each cluster in a DR result.
For our method, we adopt contrastive learning~\cite{zou2013contrastive}, a new emerging analysis approach for high-dimensional data.
Contrastive learning aims to discover ``\textit{patterns that are specific to, or enriched in, one dataset relative to another}''~\cite{abid2017contrastive}.
Among the contrastive learning methods, we specifically choose contrastive principal component analysis (cPCA)~\cite{ge2016rich,abid2017contrastive,abid2018exploring} and enhance it for visual analysis.
Our usage of cPCA, which we call ccPCA (contrasting clusters in PCA), can measure each feature's relative contribution to each cluster's \textit{contrast} to the others.
By referring to these relative contributions, users can easily focus on the features they should review in detail. 
We describe the strengths of using ccPCA with both numerical formulas and concrete examples.
In addition, because cPCA requires parameter tuning to obtain a useful result, we develop an automatic selection method that finds the best parameter value.

Moreover, we introduce a heatmap-based visualization showing all the features' contributions of each cluster. 
By employing hierarchical clustering and matrix reordering, our visualization helps the user find where clusters have similar features' contributions or how the features have similar contributions within or across clusters.
Additionally, with these methods, we are able to provide a scalable visualization that can handle the case of analyzing many features (e.g., 100 features or more).
We have built an interactive visual analytics system using ccPCA and a heatmap-based visualization.
We demonstrate the effectiveness of our methods and system with case studies using several publicly available datasets.
\section{Related Work}
\label{sec:related_work}
We survey the relevant works in (1) visualization for exploring DR results and (2) discriminant analysis and contrastive learning.

\subsection{Visualization for Exploring DR Results}

Various visualizations have been developed to assist analysis tasks for a DR result~\cite{dowling2018sirius,kwon2016axisketcher,kim2017pive,lai2018exploring,liu2014survey,liu2017visualizing}.
Here, we focus on describing the works that supports the aforementioned main task sequence (i.e., identifying clusters and understanding clusters' characteristics).
Stahnke et al.~\cite{stahnke2016probing} developed visualizations to help understand multidimensional-scaling (MDS)~\cite{torgerson1952} results.
To support a feature comparison of clusters in the MDS result, their visualization allows the user to manually select clusters and then it depicts the selected clusters' density plots for each of the features.
Similarly, for a cluster comparison in the DR results, other works~\cite{pagliosa2016understanding,marcilio2017approach,demiralp2017clustrophile,kwon2018clustervision} visualized statistical charts (e.g., bar charts and boxplots) of the features for each manually or automatically selected cluster. 
However, because the approaches in \cite{stahnke2016probing,pagliosa2016understanding,marcilio2017approach,demiralp2017clustrophile,kwon2018clustervision} depict the statistical chart for each feature, they are not scalable when there is a large number of features (e.g., 10 features). 
Broeksema et al.~\cite{broeksema2013visual} took further steps to provide a summary of the DR results. 
They developed visualizations to help understand patterns that appeared in multiple correspondence analysis (MCA)~\cite{abdi2007multiple}, which is a similar DR method as PCA for categorical data.
They visualized each data point's salient feature value extracted with MCA as a colored Voronoi cell around each projected point in the MCA result.
This linking of the DR result and the salient features helps the user interpret the DR result.
Similarly, Joia et al.~\cite{joia2015uncovering} linked the DR result and the information of features into one plot. 
In addition to an automatic selection of clusters, they obtained representative features for each cluster by using PCA. 
Afterward, they visualized these features' names as a word cloud within each clustered region instead of showing the projected points. 
Turkay et al.~\cite{turkay2012representative} also used PCA to obtain the representative features in the MDS result. 

Among the mentioned studies, the works by Joia et al.~\cite{joia2015uncovering} and Turkay et al.~\cite{turkay2012representative} are most related to ours in terms of identifying the representative features for each cluster. 
To identify such features, both methods refer to each cluster's principal components (PCs) computed by PCA (and the correlation between the features and PCs).
Even though they applied PCA within each cluster, the computed PCs might capture only the global tendency in the dataset. 
For example, all clusters may have similar or even the same PCs.
Also, their methods cannot find features that highly contribute to the differentiation or \textit{contrast} between one cluster and the others. 
It is important to provide features that make each cluster's characteristics unique.

\subsection{Discriminant Analysis and Contrastive Learning}
Discriminant\,analysis, including\,linear\,discriminant\,analysis\,(LDA)~\cite{izenman2008modern}, quadratic discriminant analysis (QDA)~\cite{mika1999fisher}, and mixture discriminant analysis (MDA)~\cite{hastie1996discriminant}, is a supervised learning method used for classification and DR.
Discriminant analysis methods use labeled data points as a learning set and construct a classifier to distinguish each class as much as possible~\cite{izenman2008modern}.
For example, LDA finds new dimensions (or components) which provide good separations between each class. 
Note that while both PCA and LDA can be categorized as linear DR methods, PCA is an unsupervised method and finds dimensions which maximize the variance of the input data points. 

As similar to PCA, we can obtain the contribution of each original dimension (or feature) to each component constructed by LDA. 
Therefore, for visual analytics, LDA has been utilized to inform the features which have an important role to distinguish clusters.
For example, Wang et al.~\cite{wang2017linear} developed linear discriminative star coordinates (LDSC). 
LDSC shows each feature's contribution to distinguishing a cluster from each other as a length of a corresponding axis of the star coordinates~\cite{kandogan2000star}. 
To obtain a better-clustered result, the user can use these axes as interfaces to discard the less contributed features or change the weight of the features used for clustering. 

While discriminant analysis is used for discriminating the data points based on their classes, contrastive learning~\cite{zou2013contrastive} focuses on finding patterns which contrast one dataset with another~\cite{abid2017contrastive}.
For example, contrastive PCA (cPCA)~\cite{ge2016rich,abid2017contrastive,abid2018exploring} is the extended version of PCA for contrastive learning. 
cPCA takes two different datasets (i.e., target and background), and then identifies the directions (or contrastive principal components) that have a higher variance in the target dataset when compared to the background dataset. 
Projection of the target dataset with these contrastive principal components provides the patterns which are uniquely found only in the target dataset. 
In addition to cPCA, several extended methods for contrastive learning have been developed (e.g., contrastive versions of latent Dirichlet allocation~\cite{zou2013contrastive}, hidden Markov models~\cite{zou2013contrastive}, regressions~\cite{ge2016rich}, multivariate singular spectrum analysis~\cite{dirie2018contrastive}, and variational autoencoders~\cite{abid2019contrastive}).

To the best of our knowledge, this paper is the first research using a contrastive learning method, specifically cPCA, for interactive visual analytics. 
We demonstrate the major advantages of using cPCA instead of PCA or LDA in \autoref{sec:find_features}.

\section{Workflow and an Analysis Example}
\label{sec:workflow_and_example}

We first define a workflow for analyzing high dimensional data using DR, and then provide an analysis example to motivate our work.  

\begin{figure}[tb]
	\centering
	\includegraphics[width=0.97\linewidth]{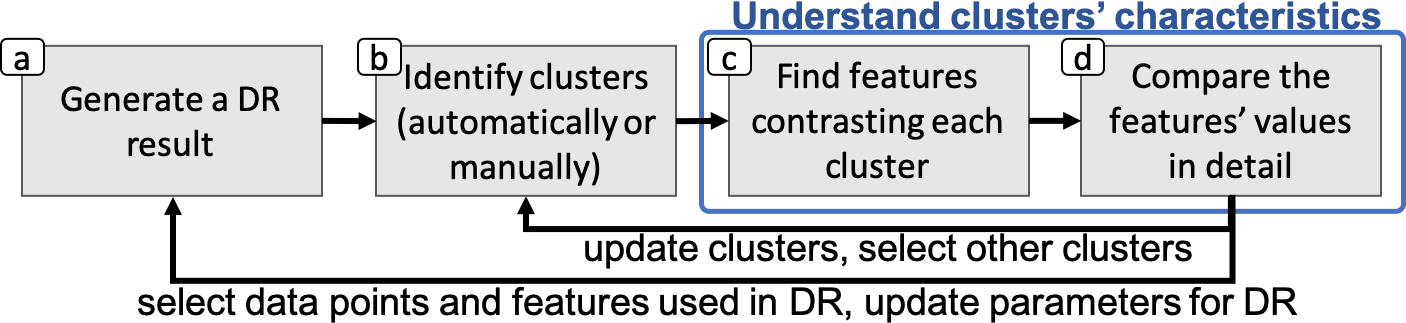}
	\vspace*{0.01in}
   	\caption{The analysis workflow.}
	\label{fig:workflow}
\end{figure}

\subsection{Analysis Workflow}
\label{sec:workflow}
\autoref{fig:workflow} shows an analysis workflow using our method. It starts from (a) applying a DR method (e.g., MDS, PCA, or t-SNE~\cite{maaten2008visualizing}) on high-dimensional data. 
Then, the task is (b) to identify clusters in the DR result by applying a clustering method (e.g., k-means~\cite{hartigan1979algorithm}, DBSCAN~\cite{ester1996density}, or spectral clustering~\cite{ng2002spectral}) or selecting clusters manually.
Afterward, the task is to understand the clusters' characteristics. 
This task has two steps. 
The first step is (c) finding features (or dimensions) which have a high contribution to contrasting each cluster with the others. 
For this step, we utilize cPCA~\cite{ge2016rich,abid2017contrastive,abid2018exploring}, as described in \autoref{sec:find_features}.
The second step is (d) reviewing the detailed differences of values of the highly contributed features between each corresponding cluster and the other data points. 
We use existing methods for DR and clustering while we introduce new methods for the last two steps.
With the last two steps, we can obtain an understanding of which and how features contribute to the uniqueness of each cluster. 
After understanding the selected clusters' characteristics, as indicated with the arrows from (d) to (a) and (b), the user can update the DR result or clusters by selecting a subset of the data points based on his/her interest, changing the parameters of the algorithms, etc.

\begin{figure*} 
\centering
\includegraphics[width=0.89\linewidth]{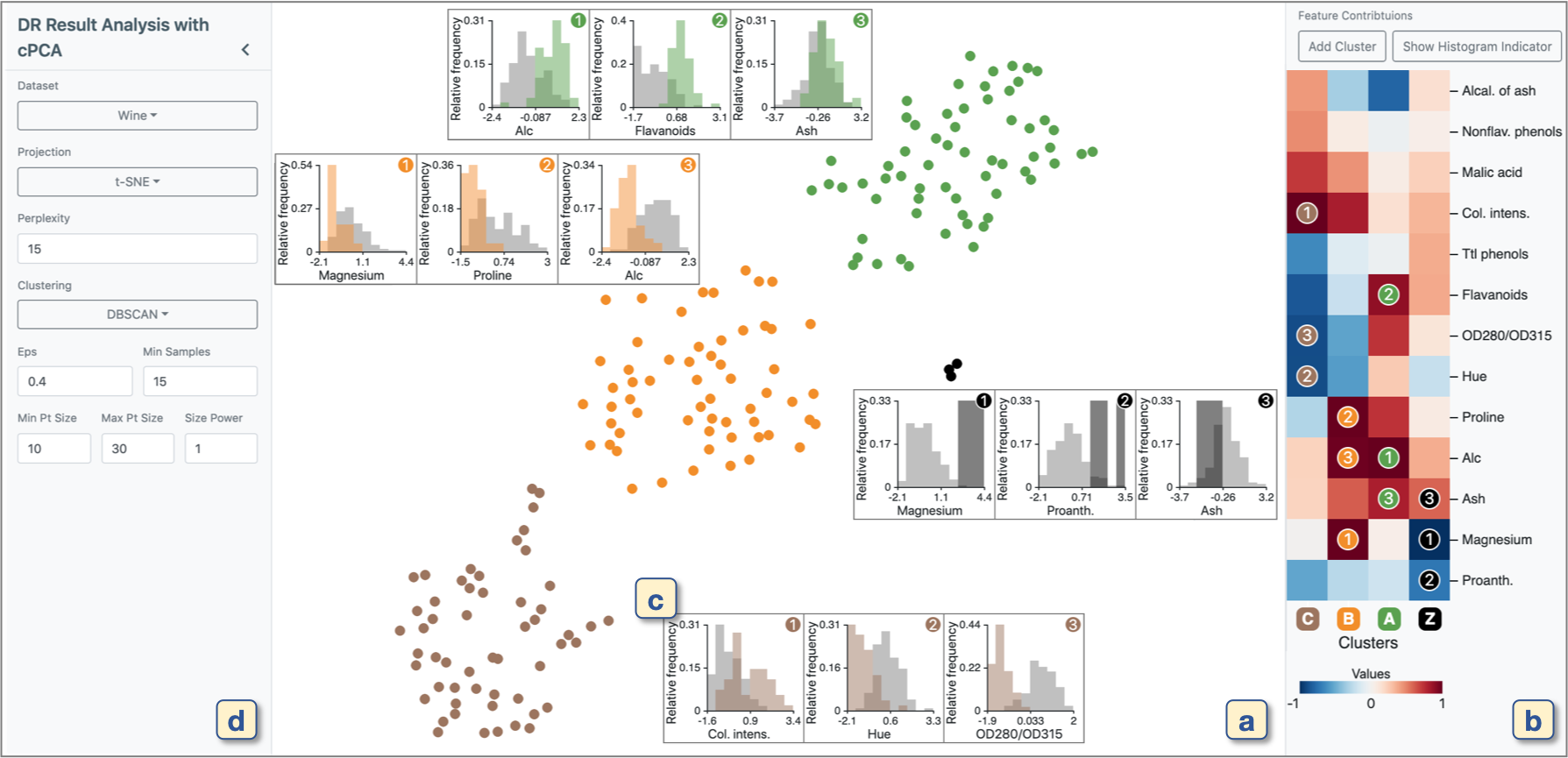}
\caption{A screenshot of our prototype system. 
The dimensionality reduction (DR) view (a) visualizes a result after DR and clustering. 
The feature contributions view (b) shows the measures of each feature's contribution to \textit{contrasting} each cluster with the others. 
The feature values of the selected cells in (b) are visualized as histograms, as shown in (c). 
In (d), we can change the settings for the analysis methods and visualizations.}
\label{fig:teaser}
\end{figure*}

\subsection{An Analysis Example}
\label{sec:example}
We analyze the Wine Recognition dataset from UCI Machine Learning Repository~\cite{uci_mlr} while following the workflow shown in \autoref{fig:workflow}.
The dataset includes 178 data points (wines) with 13 features (e.g., alcohol, color intensity, and flavanoids).  
First, to generate a DR result, we use t-SNE~\cite{maaten2008visualizing} for all of the data points. 
Then, to detect clusters, we apply DBSCAN~\cite{ester1996density} to the DR result. 
As shown in \autoref{fig:teaser}a, we identify three clusters, colored with green, orange, and brown.
The black data points are outliers or noise points labeled by DBSCAN. 
To understand the characteristics of the wines in each cluster, the system immediately applies our cPCA-based analysis method for each detected cluster. 
Now, we have obtained the features' contributions to contrasting each cluster. 
The measures of contributions are visualized with a blue-to-red divergent colormap, as indicated in \autoref{fig:teaser}b. 
As the absolute value of the measure approaches 1, the corresponding feature has a higher contribution.
Finally, for each cluster, we visualize histograms of values of the three features that have the highest contributions. 
The results are shown in \autoref{fig:teaser}c.
The histograms for each target cluster are colored with its respective cluster color, while the others are colored gray.
The $y$-axis shows relative frequency and its maximum limit is set to the maximum relative frequency of each pair of the histograms.

Based on the result shown in \autoref{fig:teaser}, we can easily perceive each cluster's characteristics. 
For example, the green cluster has higher alcohol percentage (`Alc') and flavanoids when compared to the others.
The orange cluster has lower magnesium, proline, and alcohol percentage.  
Also, the brown cluster has lower OD280/OD315 (i.e., low dilution degree), lower hue, and higher color intensity. 
The black outliers have higher magnesium and proanthocyanidins (`Proanth').

\begin{figure}[tb]
	\centering
	\includegraphics[width=0.97\linewidth,height=0.5\linewidth]{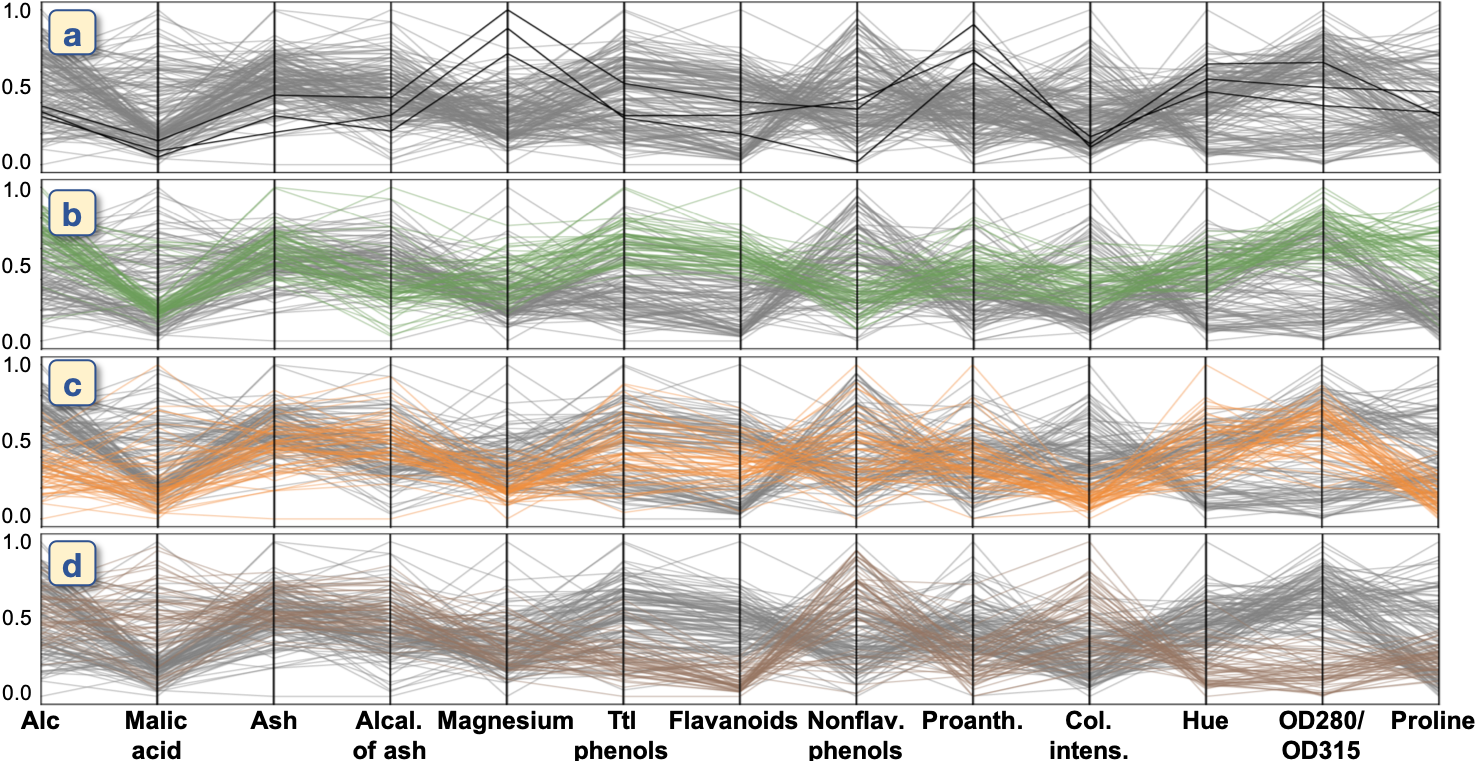}
	\vspace{1pt}
   	\caption{Parallel coordinates showing all features in the Wine Recognition dataset. The corresponding polylines for the wines are highlighted with (a) black, (b) green, (c) orange, and (d) brown clusters. It is difficult to discern the essential features from this visualization.}
	\label{fig:wine_par_coord}
\end{figure}

Even though this analysis example uses relatively a small number of features and clusters, finding these results is not a trivial task without the suggestions of highly contributed features. 
For example, in \autoref{fig:wine_par_coord}, similar to \cite{kwon2018clustervision}, we visualize each cluster's feature values with parallel coordinates~\cite{inselberg1987parallel}.
Without our method, to find the same results, the user would need to review all the features of each cluster one by one. 
This is not only time-consuming but also introduces a possibility of overlooking important characteristics.

\section{Methodology}
\label{sec:find_features}
As demonstrated in \autoref{sec:example}, when a dataset has many features, even only around ten, reviewing the values for each feature becomes tedious. 
Finding features which contrast each cluster with the other data points is the core analysis of our approach. 
To do this, we utilize cPCA~\cite{abid2017contrastive,abid2018exploring} and its linearity to obtain the features' contributions (FCs) to the contrast.

\begin{figure}[tb]
    \captionsetup{farskip=0pt}
	\centering
    \includegraphics[width=0.75\linewidth]{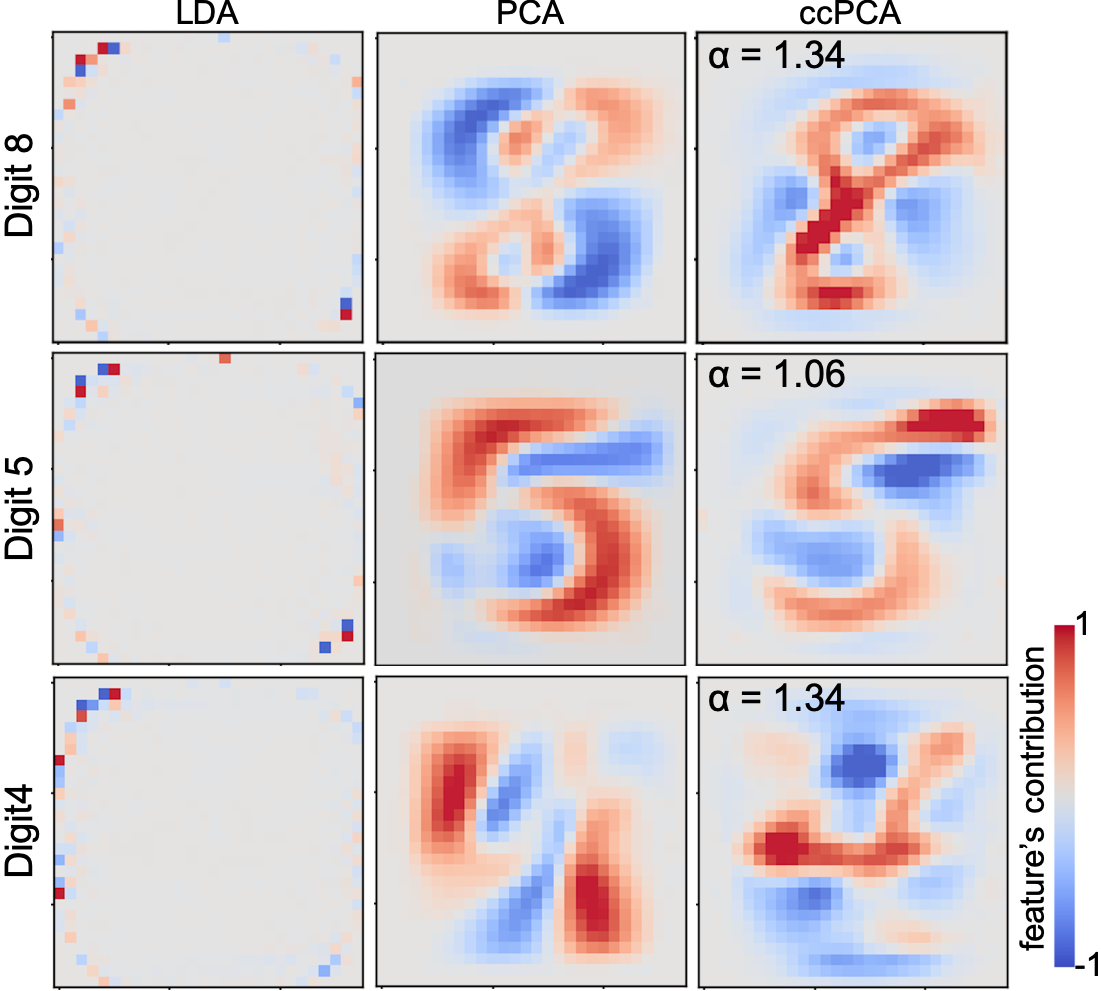}
   	\caption{
   	Comparison of features' relative contributions of MNIST digits.
   	We compare LDA, PCA, and ccPCA. 
   	All of these methods can calculate the features' relative contributions to the first component by respectively referring to either LDA's loadings, PC loadings, or cPC loadings described in \autoref{sec:feature_contributions}.
   	We scale each loading in the range from -1 to 1 by dividing the maximum absolute value of the loadings.
   	We visualize the scaled loading for each pixel with a blue-to-red colormap.
   	For LDA, we perform classification between the target digit and the others.
   	The LDA results, placed on the left column, show that the outside pixels have high contributions.
   	We can consider that LDA tries to distinguish each target digit from the others by referring to the pixels that are less frequently used in the other digits.
   	We apply PCA to each target cluster in the same manner as \cite{joia2015uncovering,turkay2012representative}.
   	We can see that the PCA results show variations of the strokes when drawing each digit.
   	The cPCA results are obtained from ccPCA with the automatic selection of $\alpha$ (refer to \autoref{sec:find_direction}).
   	When compared with PCA, the cPCA results clearly show the strokes contrasting the target digit with the others.
   	For example, for Digit 5, the pixels on the upper right have high contributions, as indicated in dark red.
   	When only drawing Digit 5, we tend to use these pixels, and thus, we can see that cPCA captures Digit 5's characteristics.
   	Similarly, for Digit 4, we can see that there are dark red pixels around the middle left.
   	}
	\label{fig:lda_pca_cpca}
\end{figure}

There is a clear advantage of using cPCA over PCA~\cite{jolliffe1986principal} and LDA~\cite{izenman2008modern}, both of which are linear DR methods.
PCA has been used to find the representative features within the selected data points~\cite{joia2015uncovering,turkay2012representative}. 
However, as shown in the examples of \autoref{fig:lda_pca_cpca}(middle), while PCA is useful to find variations within each cluster, it cannot consider the differences between one cluster and the others.
This consideration is important to find the unique characteristics in the target cluster.
On the other hand, LDA focuses only on distinguishing the target cluster from the others. 
Thus, as shown in \autoref{fig:lda_pca_cpca}(left), LDA would judge whether a feature has a high contribution to distinguishing the target cluster even in the case where the feature has little variance in the target cluster and zero variance in the others. 
This could frequently happen especially when the number of features is large.
Our cPCA-based method, ccPCA, finds the features which are well-balanced in terms of variety (similar to PCA) and separation (similar to LDA). 
Also, this balance can be controlled with the contrast parameter, as described in \autoref{sec:automatic_selection_alpha}.

\subsection{Contrastive PCA (cPCA)}
\label{sec:cpca}
We provide a brief introduction to cPCA~\cite{ge2016rich,abid2017contrastive,abid2018exploring}, which we utilize to find features contrasting a target cluster with the other data points. 
cPCA is developed for ``the setting where we have multiple datasets and are interested in discovering patterns that are specific to, or enriched in, one dataset relative to another''~\cite{abid2017contrastive}. 
For instance, from the examples in \cite{abid2017contrastive}, when we have a medical dataset $X$ of diseased patients, we would want to find trends and variations of the disease's influence.
If we apply the classical PCA~\cite{hotelling1933analysis,jolliffe1986principal} to $X$, the first principal component would only present the diseased patients' demographic variations~\cite{garte1998role}, instead of showing the variation of the disease's effects.
However, if there is another medical dataset $Y$ of healthy patients, cPCA can utilize the fact that $Y$ could have similar demographic variations as $X$, and no variations related to the disease.
By taking $X$ and $Y$ as the target and background datasets, respectively, cPCA can find the directions (or components) in which $X$ has high variance but $Y$ has low variance.

\subsubsection{Description of the Algorithm}
Now, we describe how cPCA obtains such directions by using the target and background datasets.
Let $\smash{X=\{\mathbf{x}_i\}_{i=1}^n}$ be the target dataset and $\smash{Y=\{\mathbf{y}_i\}_{i=1}^m}$ be the background dataset where $\smash{\mathbf{x}_i, \mathbf{y}_i \in \mathbb{R}^d}$, $n$ and $m$ are the numbers of data points, and $d$ is the number of dimensions (or features). 
Similar to the classical PCA, for the first step, cPCA applies centering to each dimension of $X$ and $Y$ and then obtains their corresponding empirical covariance matrices $\smash{\mathbf{C_X}}$ and $\smash{\mathbf{C_Y}}$.
Let \(\mathbf{v}\) be any unit vector of $d$ dimensions. 

\begin{figure}[tb]
    \captionsetup{farskip=0pt}
	\centering
    \includegraphics[width=0.99\linewidth]{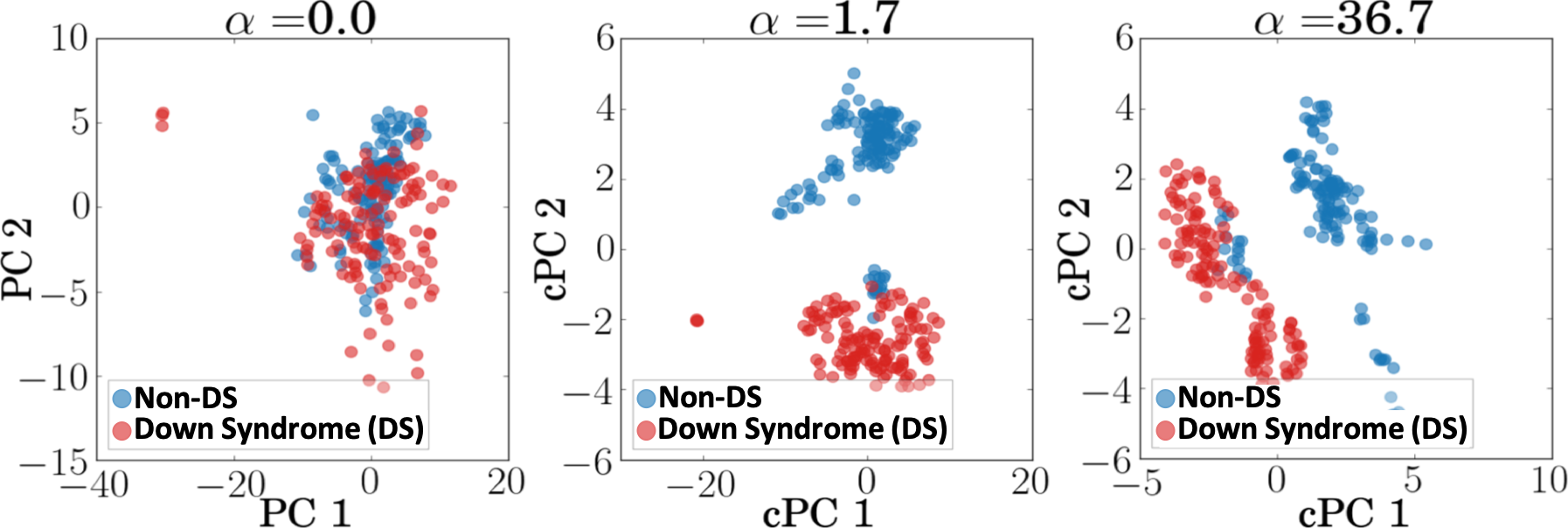}
    \vspace{1pt}
   	\caption{cPCA results of the Mice Protein Expression dataset~\cite{higuera2015self} from \cite{abid2017contrastive}. A different contrast parameter $\alpha$ value is used for each result. When $\alpha=0$, cPCA generates the same result when applying PCA to the target dataset. In this result, we cannot see clear differences between down syndrome (DS) and non-DS mice, indicated with red and blue points, respectively. While clear differences between DS and non-DS start to appear when $\alpha=1.7$, we can see that DS is further separated into two groups when $\alpha=36.7$. More examples can be found in \cite{abid2017contrastive,abid2018exploring}.}
	\label{fig:cpca}
\end{figure}

Then, with a given direction $\mathbf{v}$, the variances for the target and background datasets can be written as:
$\smash{\lambda_X(\mathbf{v}) \overset{\scriptscriptstyle\mathrm{def}}{=} \mathbf{v}^{\mathsf{T}} \mathbf{C_X} \mathbf{v}}$, $\smash{ 
\lambda_Y(\mathbf{v}) \overset{\scriptscriptstyle\mathrm{def}}{=} \mathbf{v}^{\mathsf{T}} \mathbf{C_Y} \mathbf{v}}$.
Now, the optimization that finds a direction $\smash{\mathbf{v}^*}$ where $X$ has high variance but $Y$ has low variance can be written as:
\begin{equation}
\label{eq:contrastive_direction}
     \mathbf{v}^* = \operatorname*{argmax}_{\mathbf{v}} ~ \lambda_X(\mathbf{v}) - \alpha \lambda_Y(\mathbf{v}) = \operatorname*{argmax}_{\mathbf{v}} ~ \mathbf{v}^{\mathsf{T}} (\mathbf{C_X} - \alpha \mathbf{C_Y}) \mathbf{v}
\end{equation}
where $\alpha$ is a contrast parameter $(0 \leq \alpha \leq \infty)$. We describe the details of $\alpha$ in \autoref{sec:semi_auto_selection_alpha}.
From \autoref{eq:contrastive_direction}, we can see that $\smash{\mathbf{v}^*}$ corresponds to the first eigenvector of the matrix $\smash{\mathbf{C} \overset{\scriptscriptstyle\mathrm{def}}{=} (\mathbf{C_X} - \alpha \mathbf{C_Y})}$.
The eigenvectors of \(\mathbf{C}\) can be calculated with eigenvalue decomposition (EVD). 
These computed eigenvectors are called contrastive principal components (cPCs) and are orthogonal to each other. 
Similar to the classical PCA, by using these cPCs (typically two cPCs), we can plot the DR result of $X$.
An example from \cite{abid2017contrastive} is shown in \autoref{fig:cpca}.

\subsubsection{The Contrast Parameter and Semi-Automatic Selection}
\label{sec:semi_auto_selection_alpha}
The contrast parameter $\alpha$ controls the trade-off between having high target variance and low background variance. 
When $\alpha = 0$, cPCs will only maximize the variance of the target dataset.
These cPCs are the same as the principal components (PCs) of the target dataset when computed with the classical PCA.
As $\alpha$ increases, cPCs will become more optimal directions that reduces the variance of the background dataset.
\autoref{fig:cpca} shows the example from \cite{abid2017contrastive} with different $\alpha$ values. 

As shown in \autoref{fig:cpca}, the selection of $\alpha$ has a strong impact on the DR result.
Thus, Abid and Zhang et al.~\cite{abid2017contrastive,abid2018exploring} introduced an algorithm suggesting multiple $\alpha$ values. 
Their algorithm calculates a set of cPCs for each of the multiple values of $\alpha$ (with 40 values as their default), and the $\alpha$ values are logarithmically spaced in a certain range (the default is between 0.1 and 1000). 
Then, the similarity between each pair of the different cPCs, each obtained with a different $\alpha$ value, is measured by calculating the product of the cosine of the principal angles.
Afterward, based on the user's input $p$ (the number of values of $\alpha$ to suggest), the algorithm finds $p$ clusters from the similarities with spectral clustering~\cite{ng2002spectral}.
Finally, the algorithm returns $p$ values of $\alpha$ which correspond to the medoids of the clusters. 
From the suggested $p$ values, the algorithm returns a set of DR results. 
By referring to this set, the user can choose their preferred $\alpha$ value. 

\subsection{Finding the Direction that Contrasts a Target Cluster}
\label{sec:find_direction}
As described above, cPCA discovers patterns that are specific to, or enriched in, the target dataset relative to the background dataset.
In \cite{abid2017contrastive,abid2018exploring}, cPCA is designed for the situation where the patterns the user wants to identify are included within the target dataset $X$, while the background dataset $Y$ contains the structure the user wants to remove from the target dataset. 
Therefore, in \cite{abid2017contrastive,abid2018exploring}, the provided examples for \{$X$, $Y$\} are \{`diseased subjects', `control group subjects'\}, \{`patients after treatment', `patients before treatment'\}, \{`images mixed with interests and noises', `images only including noises'\}, etc.

In our case, we want to find the directions (i.e., cPCs) which contrast one cluster with the other data points.
If we follow the examples of $X$ and $Y$ as stated above, $X$ can be the target cluster and $Y$ can be the other data points. 
However, in this case, cPCA will find cPCs that only enrich the variations specific to the target cluster. 
For example, when the target cluster includes diseased subjects and the other data points correspond to healthy subjects, cPCA will find enriched variations within the diseased subjects (e.g., differences among multiple diseases), but will not consider the differences between diseased and healthy subjects.

\begin{figure}[tb]
    \captionsetup{farskip=0pt}
    \captionsetup[subfloat]{width=0.32\linewidth}
	\centering
	\hspace*{-9pt}
	\subfloat[PCA]{
     \includegraphics[height=0.32\linewidth]{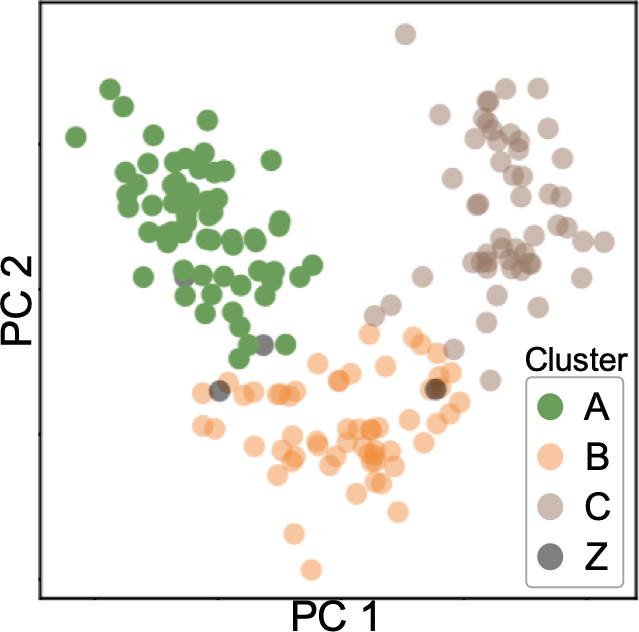}
     \label{fig:cpca_comparison_a}
    }
    \subfloat[cPCA ($\alpha=2.15$)]{
     \includegraphics[height=0.32\linewidth]{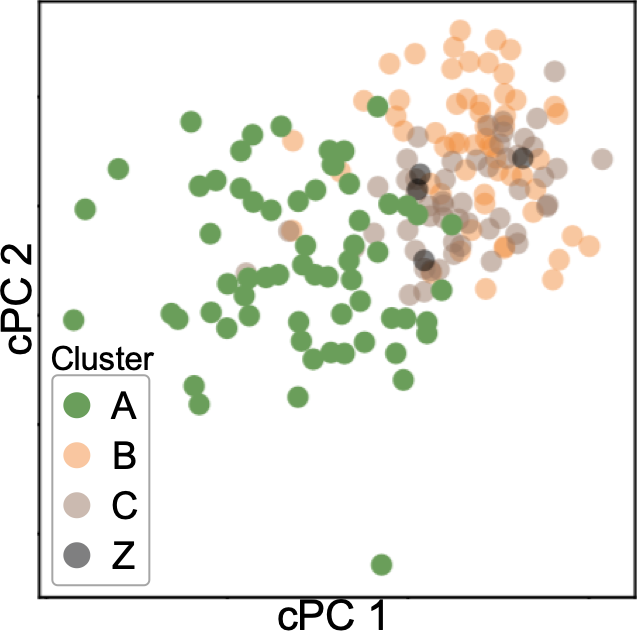}
     \label{fig:cpca_comparison_b}
    }
    \subfloat[ccPCA ($\alpha=4.38$)]{
     \includegraphics[height=0.32\linewidth]{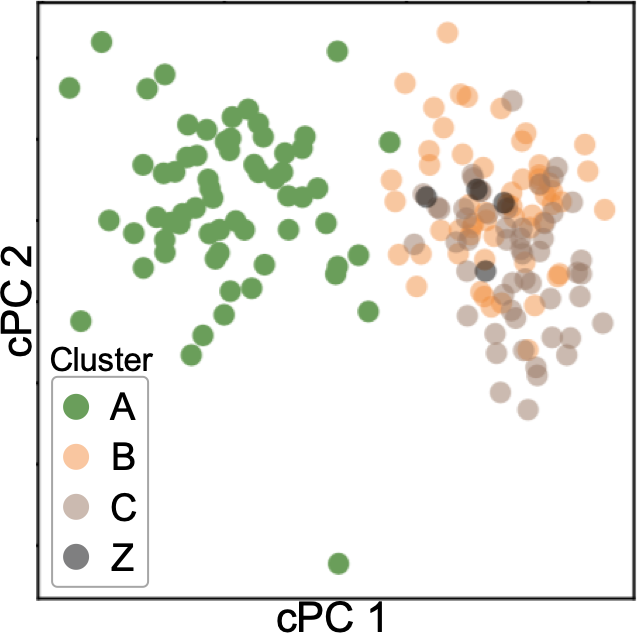}
     \label{fig:cpca_comparison_c}
    }
    \hspace*{-7pt}
    \vspace{1pt}
   	\caption{
   	The DR results of the Wine Recognition dataset. 
   	The cluster labels generated in \autoref{sec:example} are used. 
   	Here, we try to find the (c)PC contrasting the green cluster. 
   	In (a), we apply the classical PCA to the entire dataset. 
   	Though there is a separation of the green cluster when using the first and second PCs, there are overlaps of the green and orange clusters when only using the first PC (PC~1).
   	In (b), the data points in the green cluster are used as the target dataset and the other data points are used as the background dataset. 
   	$\alpha$ value is selected from the suggestions using the semi-automatic selection in \autoref{sec:semi_auto_selection_alpha}.
   	We cannot see a clear separation of the green cluster from the others.
   	In (c), we use the entire data points instead of only the green cluster as the target dataset.
   	$\alpha$ value is selected with our automatic selection method in  \autoref{sec:automatic_selection_alpha}.
   	We can see a better separation when compared to that of (a) and (b) even when using only the first cPC (cPC~1).}
	\label{fig:cpca_comparison}
\end{figure}

To utilize cPCA for finding the directions contrasting a target cluster with the others, we introduce a novel usage of cPCA, named ccPCA. 
Instead of using the target cluster as the target dataset $X$ and the other data points as the background dataset $Y$, we use the entire dataset as $X$ and the data points other than the target cluster as $Y$. 
With this approach, we can find the directions that contrast the target cluster.
As we describe in the following subsections, ccPCA has the strengths in regards to two aspects: (1) an implicit extension of the contrast parameter $\alpha$ and (2) a proper setting of the centroid. 
The DR results shown in \autoref{fig:cpca_comparison} provide a comparison of the classical PCA, original usage of cPCA (i.e., using only the target dataset as $X$), and ccPCA. 

Let $\smash{E=\{\mathbf{e}_i\}_{i=1}^{s}}$ be the entire dataset and  $\smash{K=\{\mathbf{k}_i\}_{i=1}^{t}}$ be the target cluster ($\smash{K \subset E}$, $\smash{\mathbf{e}_i, \mathbf{k}_i\in \mathbb{R}^d}$, $s$ and $t$ are the numbers of data points). 
Then, we denote $\smash{R=\{\mathbf{r}_i\}_{i=1}^{u}}$ as the difference of the two sets $K$ and $E$ (i.e., $\smash{R =  E \setminus K}$ and $u = s - t$).
With these notations, we can say that ccPCA uses $E$ and $R$ as the target $X$ and background $Y$ datasets, respectively.

\subsubsection{An Implicit Extension of the Contrast Parameter}
To provide a simple and clear explanation, we assume the centering effects to the datasets $E$, $K$, and $R$ are all the same (i.e., $E$, $K$, and $R$ have the same mean value for each feature). 
After centering the target dataset $E$ and the background dataset $R$, cPCA obtains their corresponding empirical covariance matrices $\smash{\mathbf{C_E}}$ and $\smash{\mathbf{C_R}}$. 
Then, cPCA calculates cPCs by performing EVD to $\smash{\mathbf{C_E} - \alpha \mathbf{C_R}}$. 
Let $\smash{\mathbf{C_K}}$ be the empirical covariance matrix of the target cluster $K$ after centering. 
Because $\smash{\mathbf{C_K} = \sum_{i=1}^{t}\mathbf{k}_i \mathbf{k}_i^{\mathsf{T}} / t}$, $\smash{\mathbf{C_R} = \sum_{i=1}^{u}\mathbf{r}_i \mathbf{r}_i^{\mathsf{T}} / u}$, $\smash{E = K \sqcup R}$, and $s=u+t$, $\smash{\mathbf{C_E}}$ can be represented as $\smash{\mathbf{C_E} = (t \mathbf{C_K} + u \mathbf{C_R}) / s}$. 
With this, $\smash{\mathbf{C_E} - \alpha \mathbf{C_R}}$ can be rewritten as:
\begin{align}
  \mathbf{C_E} - \alpha \mathbf{C_R} &= (t \mathbf{C_K} + u \mathbf{C_R}) / s - \alpha \mathbf{C_R} \\
                   &= \frac{t}{s} \left( \mathbf{C_K} - \frac{\left( s \alpha - u \right)}{t} \mathbf{C_R} \right)
                   = \frac{t}{s} \left( \mathbf{C_K} - \beta \mathbf{C_R} \right)
                   \label{eq:new_cpca}
\end{align}
where $\smash{\beta = (s \alpha - u) / t}$. 
Because $\smash{0 \leq \alpha \leq \infty}$, $\smash{ - u/t \leq \beta \leq \infty}$. 
Note that if we use $K$ and $R$ as the target and background datasets, respectively, cPCA performs EVD to $\smash{\mathbf{C_K} - \alpha \mathbf{C_R}}$.
Therefore, a fundamental difference between the cases of using $E$ (i.e., the entire dataset) and using $K$ (i.e., only the target cluster) as the target dataset for cPCA is the difference between $\alpha$ and $\beta$.

While $\alpha$ only takes a non-negative value, $\beta$ can be a negative value. 
When $\smash{\beta = - u/t}$, cPCA selects the directions that maximize the variance of the entire dataset $E$, and hence reduces to PCA applied on $E$. 
As $\beta$ increases to 0, cPCA provides more weight to the target cluster $K$ than the others $R$ to select the directions.
When $\smash{\beta = 0}$, cPCA selects the directions that maximize the variance of the target cluster $K$, and hence reduces to PCA applied on $K$.
Then, as $\beta$ increases from 0 to $\infty$, the directions from cPCA will become more optimal to reduce the variance of the others $R$. 
While \autoref{eq:new_cpca} with $\smash{\beta >= 0}$ has a capability to find the same directions with $\smash{\mathbf{C_K} - \alpha \mathbf{C_R}}$, ccPCA also searches the directions that considers the differences between the target cluster $K$ and the others $R$ by using the range $\smash{\beta < 0}$. 

\begin{figure}[tb]
    \captionsetup{farskip=0pt}
	\centering
	\hspace*{-7pt}
	\subfloat[Centroids and cPCs]{
     \includegraphics[width=0.35\linewidth,height=0.36\linewidth]{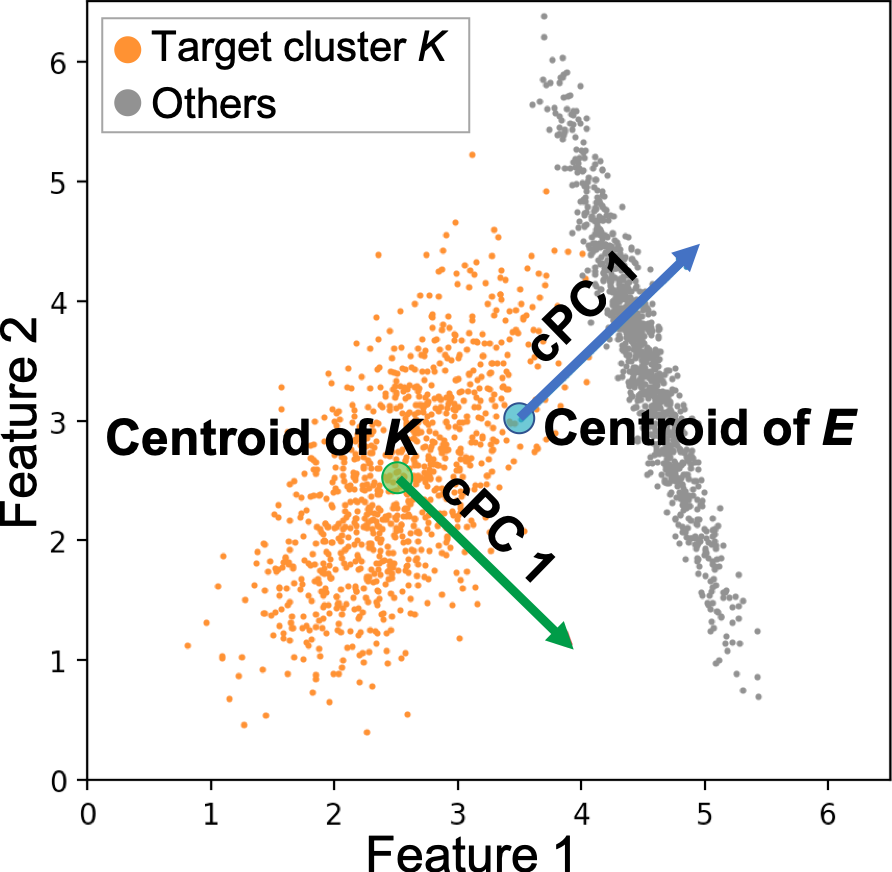}
     \label{fig:centering_a}
    }
    \hspace*{3pt}
    \subfloat[Histogram along the green cPC~1]{
     \includegraphics[width=0.28\linewidth,height=0.36\linewidth]{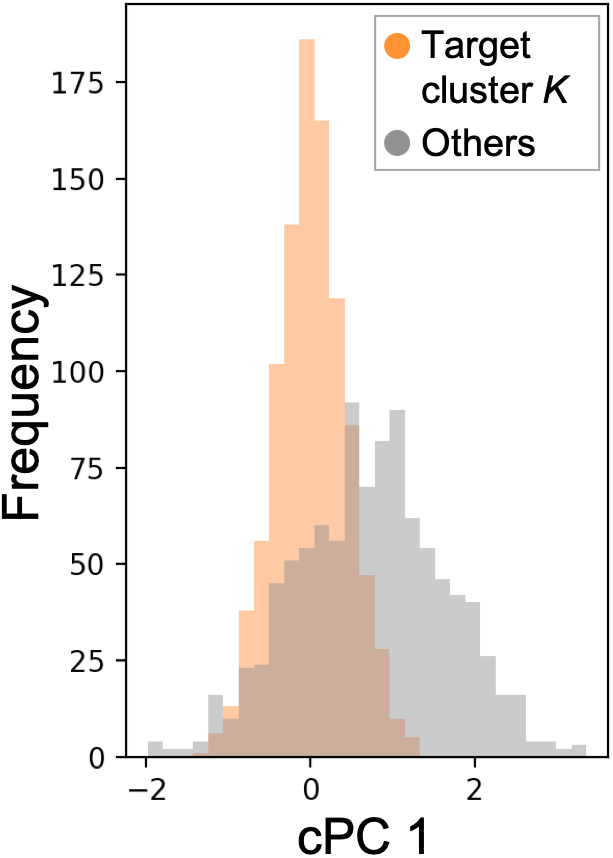}
     \label{fig:centering_b}
    }
    \hspace*{3pt}
    \subfloat[Histogram along the blue cPC~1]{
     \includegraphics[width=0.28\linewidth,height=0.36\linewidth]{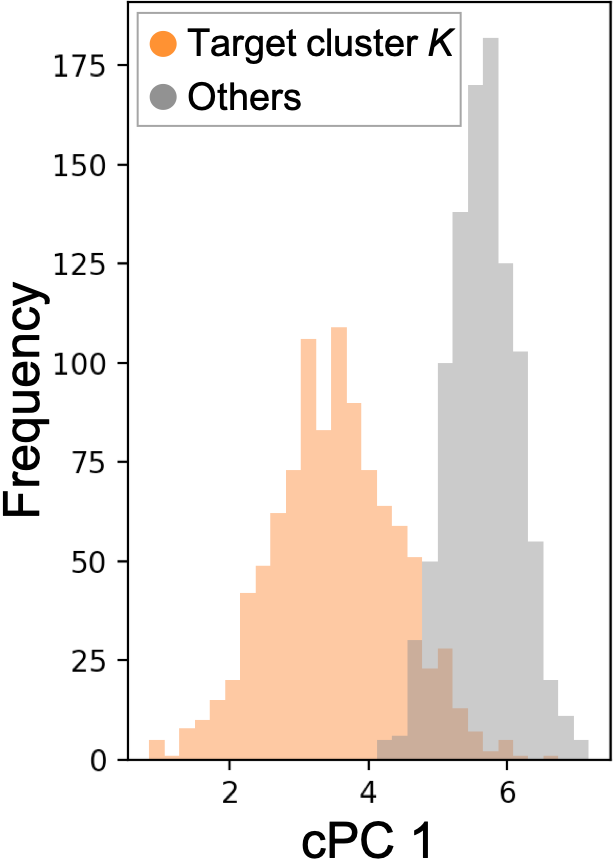}
     \label{fig:centering_c}
    }
    \hspace*{-5pt}
    \vspace{1pt}
   	\caption{
   	A comparison of centering effects to the cPCA results. 
   	For this example, we generate two sets of data points from different 2D Gaussian distributions. 
   	In (a), the green circle and arrow show the centroid and the first cPC when using the target cluster $K$ as the target dataset and the others as the background dataset. The blue circle and arrow are the centroid and the first cPC when using the entire dataset $E$ as the target dataset.
   	From the DR results, as shown in (b) and (c), ccPCA, using the entire dataset $E$ as the target dataset, generates a better separation between the target cluster and the others.}
	\label{fig:centering}
\end{figure}

\subsubsection{The Centering of the Target Dataset}
ccPCA not only implicitly extends the searching range of $\alpha$ of $\smash{\mathbf{C_K} - \alpha \mathbf{C_R}}$, but it also uses a proper centroid of the dataset.
The centering (i.e., the mean subtraction for each feature) in cPCA is used for translating the dataset to its centroid. 
When using $K$ as the target dataset, the centroid is calculated from only the target cluster $K$.
In contrast, ccPCA uses $E$ as the target dataset, and the centroid is calculated from all the data points.
\autoref{fig:centering} shows an example of the two methods of calculating the centroid and the first cPC in each case. 
As the same reason as the classical PCA, the centering should be applied to the entire dataset in our case. 
This is to ensure that the first cPC is the direction of the maximum variance, which contrasts the differences between the target cluster and the others.

\subsubsection{Automatic Selection of the Best Contrast Parameter}
\label{sec:automatic_selection_alpha}
\begin{figure}[tb]
    \captionsetup{farskip=0pt}
	\centering
    \includegraphics[width=1.0\linewidth]{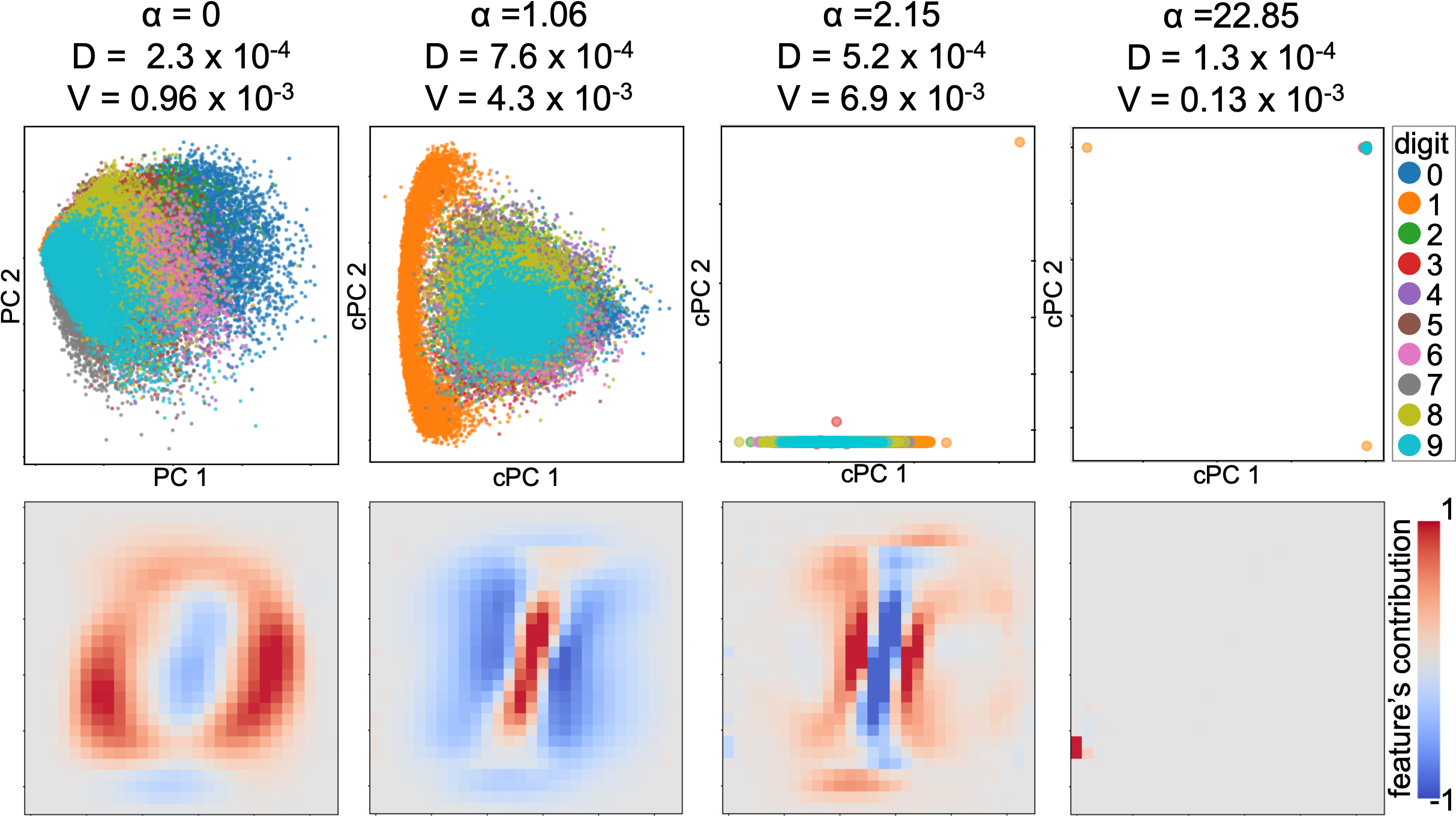}
   	\caption{
   	The DR results with the first and second cPCs (top row) and the features' (i.e., pixels') relative contributions (bottom row) of the MNIST dataset~\cite{mnist} with different $\alpha$ values (refer to \autoref{sec:feature_contributions} about the features' relative contributions).
   	Here, we try to contrast Digit 1 with the other digits.
   	We can see that when $\alpha=0$ (reduced to applying the classical PCA for all digits), cPCA does not separate Digit 1, and the features' contributions do not show any useful information to understand the characteristics of Digit 1. 
   	On the other hand, when $\alpha=22.85$, while some of Digit 1 (e.g., points placed on the top left) are well separated, the variance $V$ is small. 
   	Also, from the features' contributions, we can see that only a few pixels in the lower left have high contributions.
   	This is expected because these pixels are rarely used when drawing Digit 1.
   	The result with $\alpha=1.06$ produces the best discrepancy score $D$ and a large variance $V$.
   	This $\alpha$ will be selected by \autoref{eq:var}.
   	Also, we can see that cPCA highlights the pixels around the center, which are typically used for drawing Digit 1.
   	}
	\label{fig:auto_selection_alpha}
\end{figure}

The selection of the contrast parameter $\alpha$ is the remaining procedure.
Even though we can use the existing semi-automatic selection of $\alpha$ in \autoref{sec:semi_auto_selection_alpha}, selecting the best alpha from the multiple suggested options is tedious when analyzing multiple clusters. 
Thus, we introduce a method for an automatic selection of the best $\alpha$ for our usage.
The pseudocode of this method is available in the Supplementary Materials~\cite{supp}.
To understand the characteristics of the cluster, we should find the first cPC which not only (1) shows a clear separation between the target cluster from the others, but also (2) maintains the variability in the target cluster well (i.e., a high variance within the target cluster). 
Similar to the classical PCA, the second condition tries to preserve the target clusters' original structure.
Without the second condition, when using a large $\alpha$, cPCA may preferentially select features where the target cluster only has subtle variability, but the other data points have no variability (i.e., zero variance). 
This example can be seen in the far right of \autoref{fig:auto_selection_alpha}.

Similar to the semi-automatic selection in \autoref{sec:semi_auto_selection_alpha}, our automatic selection lists multiple candidates of $\alpha$ (our default is also 40 values).
These candidates consist of 0 and a set of logarithmically spaced values given a certain range (our default also ranges from 0.1 to 1000).
We denote these alphas as \(\{\alpha_i\}_{i=1}^q\) ($q$ is the number of candidate values for the best $\alpha$) and assume \(\{\alpha_i\}\) is sorted by ascending order (i.e., \(\alpha_1 = 0\)). 
Then our method selects a value that obtains the best separation while having enough variance in the target cluster $K$.

To measure the separation between the target cluster and the others along the first cPC, we use the histogram intersection (HI)~\cite{swain1991color}, which can measure the overlaps of the histograms of the two sets. 
While there are many different (dis)similarity measures between two probability distributions, such as the Kullback-Leibler divergence~\cite{kullback1951information}, we chose HI for its robustness to outliers and low computational cost.
Let $H_A = \{{h_A}_{\scriptscriptstyle j}\}_{j=1}^{b}$, $H_B = \{{h_B}_{\scriptscriptstyle j}\}_{j=1}^{b}$ be the histograms of two given sets of real numbers $A$ and $B$ where $b$ is the number of bins, $\smash{{h_A}_{\scriptscriptstyle j}}$ and $\smash{{h_B}_{\scriptscriptstyle j}}$ are the numbers of data points in the $j$-th bin of $A$ and $B$, respectively. 
Both $\smash{H_A}$ and $\smash{H_B}$ have the same bins.
We decide the bin-width using Scott's normal reference rule~\cite{scott1979optimal} from the set of real numbers obtained by combining $A$ and $B$.
The HI of the two sets $A$ and $B$ is defined as: \(I(A, B) = \sum_{j=1}^b \min({h_A}_{\scriptscriptstyle j}, {h_B}_{\scriptscriptstyle j})\).
Let \(K'_i\) and \(R'_i\) be the data points of 1D DR results of $K$ and $R$ with the first cPC corresponding to the $i$-th candidate $\alpha$ value (i.e., $\alpha_i$), respectively.
Then, we can calculate the measurement of separation with the inverse HI (i.e., $\smash{I(K_i', R_i')^{-1}}$) for each $\alpha_i$. 
We refer $\smash{I(K_i', R_i')^{-1}}$ as the discrepancy score $\smash{D(\alpha_i)}$.

For the variance of \(K_i'\), to handle the scaling differences in each DR result, first, we apply the min-max scaling to \(K_i'\) with the minimum and maximum values of $K_i' \sqcup R_i' $. 
Then, we calculate the variance of the scaled \(K_i'\). 
We denote this variance of \(K_i'\) as \(V(\alpha_i)\).

With the measures of \(D(\alpha_i)\) and \(V(\alpha_i)\), our automatic selection method selects the best alpha with:
\begin{equation}
    \label{eq:var}
     \operatorname*{argmax}_{\alpha_i \in \{\alpha_1, \ldots, \alpha_q\}} D(\alpha_i) \ \ s.t. \ \  V(\alpha_i) \geq \gamma V(\alpha_1)
\end{equation}
where $\gamma \ (\gamma \geq 0)$ is a ratio that controls the threshold of the variance \(V(\alpha_i)\). 
Note that \(V(\alpha_1)\) is the variance of \(K'_1\) of the cPCA result with $\alpha = 0$, which will be the same result when applying the classical PCA to the entire dataset $E$.
While our method allows the user to select any non-negative value for $\gamma$, we set $\gamma = 0.5$ as the default to ensure that \(V(\alpha_i)\) has at least a half of \(V(\alpha_1)\). 
\autoref{fig:auto_selection_alpha} shows the cPCA results with different $\alpha$ values. 
Our automatic $\alpha$ selection chooses $\alpha=1.06$ in this case. 
More comprehensive experimental results with various datasets and $\alpha$ values can be found in the Supplementary Materials~\cite{supp}.

In summary, the original cPCA is enhanced as ccPCA by using \autoref{eq:contrastive_direction} with $X = E$ and $Y = E \setminus K$ and by selecting $\alpha$ as the solution to \autoref{eq:var}.

\vspace{2pt}
\noindent\textbf{Parallel calculation of the best contrast parameter:} The original semi-automatic selection of the contrast parameter in \cite{abid2017contrastive,abid2018exploring} calculates cPCA for each \(\alpha_i \in \{\alpha_i\}_{i=1}^q\) in serial~\cite{original_cpca} ($q=40$ by default).
Because the calculation of cPCA for each \(\alpha_i\) is independent of each other, in order to achieve faster computation, our method uses multi-threads and calculates each cPCA result, \(D(\alpha_i)\), and \(V(\alpha_i)\) in parallel. 
The comparison of the completion time of the original cPCA and our implementation with and without parallelization is available in \cite{supp}.

\subsection{Features' Relative Contributions to the First cPC}
\label{sec:feature_contributions}
By using cPCA, with our automatically selected $\alpha$, we can now obtain the direction (i.e., the first cPC) that contrasts the target cluster. 
Next, we determine how strongly each feature of the target cluster contributes to this direction. 
Similar to the classical PCA, by using the top eigenvalue $\smash{\lambda^*}$ and the corresponding eigenvector $\smash{\mathbf{v}^*}$ (i.e., the first cPC) of the matrix $\mathbf{C_E} - \alpha \mathbf{C_R}$, the relative contributions can be calculated with: $\mathbf{w}^* = \sqrt{\lambda^*} \mathbf{v}^*$ where $\smash{\mathbf{w}^*=\{w_i^*\}_{i=1}^d}$ ($\smash{-1 \leq w_i^* \leq 1}$).
Analogous to the classical PCA, we call $\smash{\mathbf{w}^*}$ the cPC loadings of the first cPC. 
As $\smash{|w_i|}$ approaches 1, the $i$-th feature has a stronger contribution (or correlation) to the first cPC.
Based on this value, we can decide which features we should review to understand the target cluster. 
\autoref{fig:lda_pca_cpca} shows an example of the features' contributions and comparisons with the results from LDA and PCA.
Comprehensive comparisons of LDA and PCA, using multiple datasets, can be found in the Supplementary Materials~\cite{supp}.
As shown in \autoref{fig:lda_pca_cpca}, signed cPC loadings can clearly differentiate features whose positive centered values contribute to the negative or positive direction of the first cPC by using blue and red, respectively. This is as opposed to taking the absolute value of the signed cPC loadings. 

\section{Visual Analytics System}
To demonstrate our methods of analyzing real-world datasets, we develop a prototype system that supports the analysis workflow shown in \autoref{fig:workflow}.
A major portion of the system's functionality and a video of an interaction demonstration are available in our online site~\cite{supp}.

\subsection{Dimensionality Reduction View}
The dimensionality reduction (DR) view, as shown in \autoref{fig:teaser}a, is used for the first two processes: generating a DR result and identifying clusters. 
In this view, first, the user can visualize a 2D DR result of a high-dimensional dataset. 
We employ t-SNE~\cite{maaten2008visualizing} (specifically, Barnes-Hut t-SNE implementation~\cite{van2014accelerating}) as a DR method because t-SNE can effectively depict the local structure of the dataset, and thus, it is useful to visually identify the clusters within the dataset.
From the settings in \autoref{fig:teaser}d, the user can adjust the perplexity parameter of t-SNE, which controls a balance of the effects from local and global structures of the dataset~\cite{maaten2008visualizing}.
While a larger perplexity will preserve more of the distance relationship in the global structure, a smaller perplexity will focus on more preserving the distance relationship among a small number of neighbors.

After obtaining the DR result with t-SNE, the user can identify clusters automatically or manually. 
As a default, the automatic clustering method will be immediately applied to the obtained DR result. 
As part of the automatic method, our system supports DBSCAN~\cite{ester1996density} because the density-based clustering algorithm is able to identify clusters with arbitrary shapes~\cite{parimala2011survey}, which are often generated from DR. 
The user can change the parameters required for DBSCAN from the settings in \autoref{fig:teaser}d. 
The categorical color of each point in the DR result is assigned to the clustering label obtained from DBSCAN.
The color black, in particular, is used to represent outliers or noise points labeled by DBSCAN. 
For a manual selection of a cluster, the system supports a rectangle selection. 
The user can select data points by drawing a rectangle with mouse dragging in the DR result.
Also, the user can add additional data points or unselect data points by using different selection modes provided in the system.
From these interactions, the user can create a new cluster consisting of the selected points by clicking the ``Add Cluster'' button placed at the top of \autoref{fig:teaser}b. 
The system also supports basic view-level interactions, such as zooming and panning. 

\subsection{Features' Contributions View}
\label{sec:features_contributions_view}
The two remaining processes (i.e., finding features contrasting each cluster and comparing the features' values in detail) are performed with the features' contributions (FCs) view shown in \autoref{fig:teaser}b.
In the FCs view, the FCs contrasting each cluster described in \autoref{sec:feature_contributions} are visualized as a heatmap.
While each row name shows the corresponding feature, each column name shows the cluster label (`Z' is used to represent the outliers, noise points, or both). 
Also, to indicate the corresponding cluster in the DR view, the background of each column name is colored with the corresponding color.
We scale each cluster's FCs in the range from $-1$ to $1$ by dividing each FC by the maximum absolute value of the FCs (e.g., the original range from $-0.1$ to $0.5$ will be changed to the range from $-0.2$ to $1.0$).
Then, we encode the scaled FCs with a blue-to-red colormap. 
In the next subsections, we describe our algorithm organizing the heatmap.

\begin{figure}[tb]
    \captionsetup{farskip=0pt}
	\centering
    \includegraphics[width=1.0\linewidth]{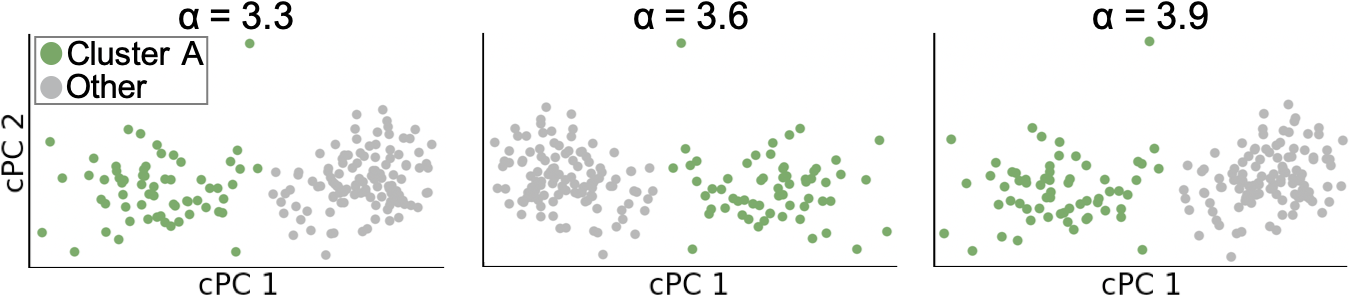}
   	\caption{
   	Sign flipping of cPCs. 
   	We generate the cPCA results of the Wine Recognition dataset used in \autoref{sec:example} with different $\alpha$ values.
   	Sign flipping occurs between $\alpha=3.3$ and $\alpha=3.6$; $\alpha=3.6$ and $\alpha=3.9$. 
   	}
	\label{fig:sign_flip}
\end{figure}

\subsubsection{Optimal Sign Flipping of cPCs and FCs}
Similar to the classical PCA, cPCA has the ``sign ambiguity'' problem~\cite{bro2008resolving,jeong2009understanding,fujiwara2019incremental}.
Because of this problem, arbitrary sign flipping in each (c)PC occurs when performing EVD. 
An example of sign flipping in cPCA is shown in \autoref{fig:sign_flip}.
Sign ambiguity affects the comparison of the FCs among the clusters.
Each cluster might have the opposite direction of the first cPC only due to this sign ambiguity problem. 
In this case, the FCs also have opposite signs, and thus, it is difficult to judge whether these clusters have similar patterns in the FCs or not.

To solve this problem as much as possible, we introduce a method to optimally reduce unnecessary sign flipping. 
Let $\mathbf{v}^*_i$ and $\mathbf{v}^*_j$ be the first cPCs of $i$-th and $j$-th clusters, respectively.
We can measure how the directions $\mathbf{v}^*_i$ and $\mathbf{v}^*_j$ are similar with the cosine similarity $sim(i, j) = \mathbf{v}^*_i \cdot \mathbf{v}^*_j / (\|\mathbf{v}^*_i\| \|\mathbf{v}^*_j\|)$.
$\mathbf{v}^*_i$ and $\mathbf{v}^*_j$ have the same direction when $sim(i, j) = 1$, while $\mathbf{v}^*_i$ and $\mathbf{v}^*_j$ have opposite directions when $sim(i, j) = -1$.
Ideally, by flipping the signs of the first cPCs of some clusters, we want to ensure that all of the clusters' first cPCs face the same side (i.e., $sim(i,j) \geq 0\ \ \forall i, j$). 
However, the sign flipping to a certain cluster affects all cosine similarities related to this particular cluster.
Thus, in many cases, it is theoretically impossible to obtain the result stated above.
However, alternatively, we can maximize the sum of all $sim(i, j)$ with sign flipping. 
This optimization can be written as:
\begin{eqnarray}
     \operatorname*{argmax}_{\varphi = \{\varphi_i, \ldots, \varphi_l\}} \sum_{i=1}^{l} \sum_{j=1, j \neq i}^{l} \frac{ (\varphi_i \mathbf{v}^*_i) \cdot (\varphi_j \mathbf{v}^*_j)}{\|\mathbf{v}^*_i\| \|\mathbf{v}^*_j\|} = \sum_{i=1}^{l} \sum_{j=1, j \neq i}^{l} \varphi_i \varphi_j sim(i, j) \nonumber \\ 
     s.t.~\varphi_i, \varphi_j \in \{-1, 1\}
     \label{eq:sign_flip}
\end{eqnarray}
where $l$ is the number of clusters and $\varphi$ is a set of signs.

We solve \autoref{eq:sign_flip} with a heuristic approach.
We initialize $\varphi=\{1, 1, \ldots, 1\}$.
We can expect that there is a higher chance to obtain a better result if we start to flip the sign where
$i$-th cluster has the largest negative value in the sum of the similarities ($\sum_{j=1}^l \varphi_i \varphi_j sim(i, j)$).
Therefore, our approach first checks whether sign flipping to the first cPC of such a cluster provides a better result in the objective function of \autoref{eq:sign_flip}. 
If so, we flip its first cPC's sign.
Then, we repeatedly apply this procedure until $\sum_{j=1}^l \varphi_i \varphi_j sim(i, j) \geq 0$ for all $i \in \{1, 2, \ldots, l\}$ is satisfied or all clusters have been checked.
Afterward, based on the optimized set $\varphi$, we allocate the new signs to respective cPC and FCs for each cluster.

\subsubsection{Ordering of Features and Clusters}
The FCs view can be used for finding not only the heatmap cells which have high FCs, but also the clusters which have similar FC patterns; the features which have similar FCs within and/or among clusters.
The case when the clusters have similar FCs implies that these clusters are contrasted due to the same features, but they have different distributions in their features' values. 
When the features have similar FCs, by reviewing the distributions of one of these features' values, we can expect that the other features may also have similar distributions. 

\begin{figure}[tb]
    \captionsetup{farskip=0pt}
	\centering
	\hspace*{-4pt}
	\subfloat[Original]{
     \includegraphics[height=0.56\linewidth]{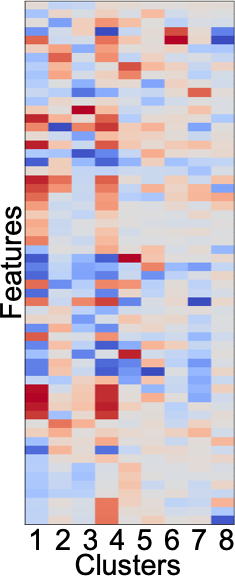}
     \label{fig:reordering_a}
    }
    \hspace*{4pt}
    \subfloat[Reordered]{
     \includegraphics[height=0.56\linewidth]{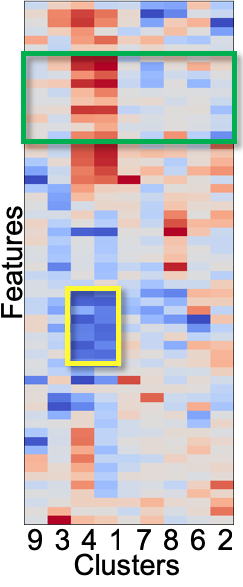}
     \label{fig:reordering_b}
    }
    \hspace*{2pt}
    \subfloat[Aggregated]{
     \includegraphics[height=0.56\linewidth]{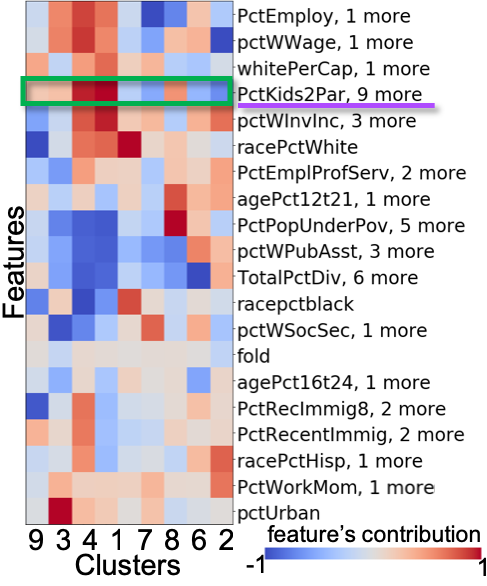}
     \label{fig:reordering_c}
    }
    \vspace{2pt}
   	\caption{
   	Reordering and aggregation of the FCs. (a) shows the original FCs. 
   	There are 8 clusters and 60 features. 
   	(b) shows the reordered FCs in both rows (i.e., features) and columns (i.e., clusters). 
   	With (b), we can see a group of similar FCs (e.g., the features are indicated with a yellow rectangle).
   	In (c), the 60 feature clusters are aggregated into 20 rows. 
   	For example, the ten features indicated with the green rectangle in (b) is aggregated into one row indicated with the green rectangle in (c).}
	\label{fig:reordering}
\end{figure}

To help find these patterns, our system applies reordering of the features (i.e, rows) and clusters (i.e., columns) based on the FCs.
Ordering choice is important since this affects how easily we can find patterns in a heatmap~\cite{behrisch2016matrix}.
We use a hierarchical clustering, specifically the complete-linkage method~\cite{mullner2011modern}, with the optimal-leaf-ordering~\cite{bar2001fast}.
Recent survey~\cite{behrisch2016matrix} reported that this combination tends to produce a coherent and quality result to help reveal patterns. 
\autoref{fig:reordering_a} and b show the results before and after the reordering. 
From \autoref{fig:reordering_b}, we can easily see a group of similar FCs.

\subsubsection{Scalable Visualization}
\label{sec:aggregation}
When the number of features is large (e.g., 100 or more), the heatmap-based visualization would have a scalability issue. 
Moreover, in this case, many features could have high FCs, and as a result, it would still be difficult to decide which features we should review in detail. 
To solve this issue, we introduce an aggregation method, utilizing the hierarchical clustering result obtained through the reordering method. 

When the number of features is larger than threshold $\delta$ (we set $\delta=40$ as a default), our method obtains $\delta$ clusters from the features by referring to the hierarchical clustering result. 
Then, our method aggregates the FCs into one representative value: the mean or the maximum absolute value.
As a default, our method takes the maximum absolute value to show the most prominent feature.
\autoref{fig:reordering_c} shows an example of the aggregation.
Additionally, to provide a representative name for each aggregated feature, our method chooses the name based on which FC has the maximum absolute value. 
With this name, our method also shows how many features are aggregated in each row, as shown with a purple underline on the right side of \autoref{fig:reordering_c} (`PctKids2Par, \textit{9 more}').

\subsection{Interactions between Views}
\textbf{From DR View:}
When the user updates the clusters with the clustering method in the DR view, the FCs view updates the heatmap with the reordering (and aggregation) method(s). 
When the user adds a new cluster manually, the\,FCs\,view\,updates\,the\,heatmap\,with\,the\,new\,cluster.

\vspace{3pt}
\noindent\textbf{From FCs View:}
The FCs view can be used as an interface to compare the details of the features' values within/across features or clusters.
When the user places the mouse over a certain heatmap cell, the system shows a popup window of the histograms of feature values of the corresponding cluster and the others (e.g., \autoref{fig:teaser}c and \autoref{fig:crime_fc}). 
We color the selected cluster's histogram with a categorical color representing its cluster label, while the gray color is used for the other data points' histogram. 
When hovering over a certain (representative) feature name, the system shows a value of the (representative) feature as the size of each data point in the DR view (e.g., \autoref{fig:nutrients_1}a and \autoref{fig:nutrients_2}a).  

Moreover, when hovering over a certain cluster label, the system highlights the corresponding cluster in the DR view.
In addition, with the popup window, the system visualizes the histograms of 1D DR results of the cluster and the others. 
From these histograms, the user can grasp how well the cluster is contrasted with the other data points. 
Additionally, the system shows the histograms of three (representative) feature values that have the highest absolute FCs.
These histograms are useful to understand each cluster's characteristics quickly.

Also, to make the comparison within/across features or clusters easier, our system allows the user to prevent the histograms from disappearing with a mouse-click.
The clicked histograms can also be moved with mouse-dragging. 
The corresponding heatmap cell for each histogram is annotated with a gray line and a pair of numbers shown in the heatmap cell and the histogram (e.g., \autoref{fig:teaser} and \autoref{fig:crime_fc}). 
The gray line can be turned on or off by clicking the ``Show/Hide Histogram Indicator'' placed at the top of the FCs view. 

\subsection{Implementation}
We have developed our system as a web application. 
To achieve fast calculation, we have implemented our methods described in \autoref{sec:find_features} with C++ and Eigen library~\cite{eigen} for linear algebraic calculations. 
We have also provided Python bindings for our C++ implementation.
The source code is available in \cite{supp}.
The back-end of the system uses Python with the stated bindings.
The front-end visualization is implemented with a combination of Elm~\cite{elm}, HTML5, JavaScript, WebGL, and D3~\cite{bostock2011d3}. 
While we use D3 for the FCs view, WebGL is used to render the data points efficiently for the DR view.
We use WebSocket to communicate between the front- and back-ends.

\begin{figure}[tb]
    \captionsetup{farskip=0pt}
	\centering
    \includegraphics[width=1.0\linewidth]{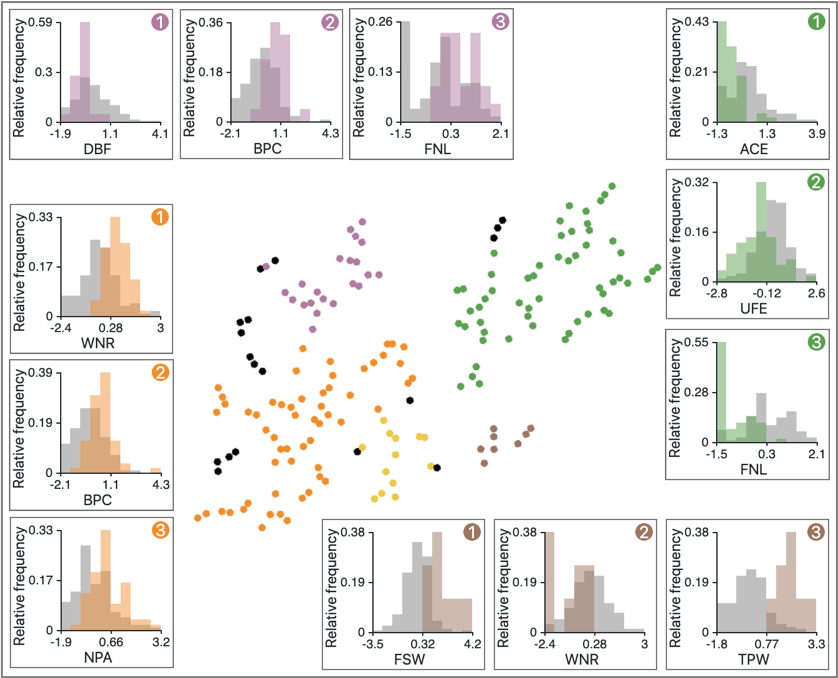}
  	\caption{An analysis result of female players from the Tennis Major Tournament Match Statistics dataset.}
	\label{fig:tennis_women}
\end{figure}

\section{Case Studies}
We have shown the effectiveness of our methods with the Wine Recognition~\cite{uci_mlr} and MNIST~\cite{mnist} datasets in the previous sections.
We demonstrate three additional case studies with publicly available datasets. 
For each case study, we preprocess the corresponding dataset to clean up missing values in the data or extract useful information for the analysis.
All the preprocessed datasets are available online~\cite{supp}.

\subsection{Tennis Major Tournament Match Statistics}
We analyze the Tennis Major Tournament Match Statistics dataset from UCI Machine Learning Repository~\cite{uci_mlr}.
This dataset contains the match statistics for both females and males at four major tennis tournaments in 2013. 
The statistics include first serve won by each player, double faults committed by each player, etc.
From this dataset, we obtain female players' mean values for each statistic across all tournaments. 
The obtained dataset consists of 174 data points (tennis players) and 13 features (statistics).

Similar to the analysis of \autoref{sec:example}, we obtain the DR result with t-SNE, clusters with DBSCAN, and FCs with our methods. 
Then, to analyze each cluster's characteristics, we show the histograms of the top 3 contributed features.
The result is shown in \autoref{fig:tennis_women}.

From \autoref{fig:tennis_women}, we can see that each cluster has a different playing style. 
For example, the purple cluster tends to have low `DBF' (double faults committed by player), high `BPC' (break points created by player), and high `FNL' (final number of games won by player). 
This indicates that these players had fewer mistakes in their serves and performed well when they were the receiver, and as a result, they won more games.
Similarly, the orange cluster has high `WNR' (winners earned by player) and `NPA' (net points attempted by player). 
These statistics will tend to be higher when a player tries to obtain points aggressively during a rally. 
On the other hand, the brown cluster has low `WNR' but high `FSW' (first serve won by player) and `TPW' (total points won by player). 
Therefore, we can say that these players tend to obtain more points with their serves.

\begin{figure}[tb]
    \captionsetup{farskip=0pt}
	\centering
    \includegraphics[width=1.0\linewidth]{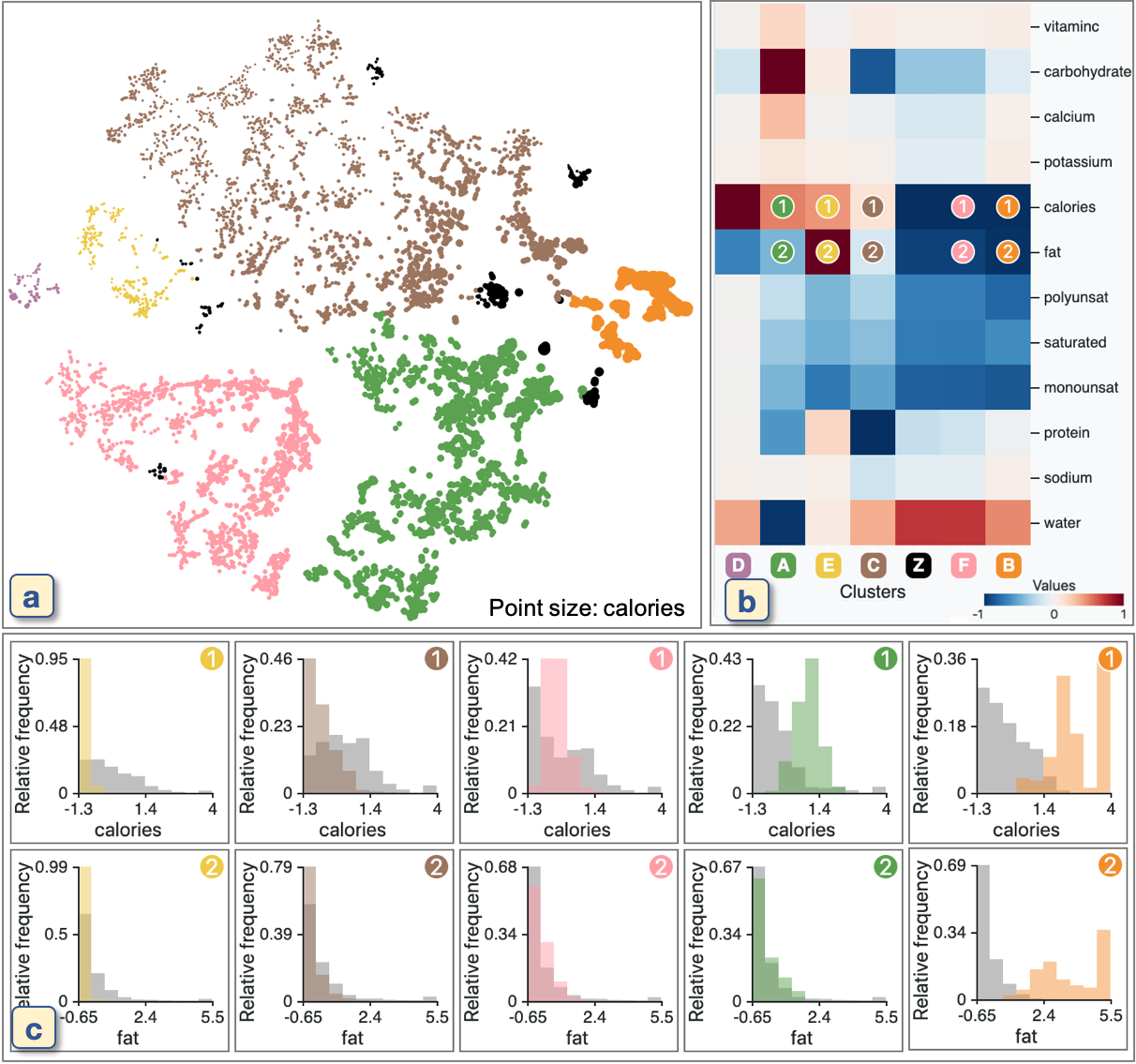}
   	\caption{A result of the Nutrient dataset. (a) shows the result after applying t-SNE and DBSCAN. A point's color and size show the clustering label and the value of `calories', respectively. (b) shows the FCs of each cluster. (c) shows the histograms of the selected cells in (b), as indicated with the colored numbers in both (b) and (c).}
	\label{fig:nutrients_1}
\end{figure}

\begin{figure}[tb]
    \captionsetup{farskip=0pt}
	\centering
    \includegraphics[width=1.0\linewidth]{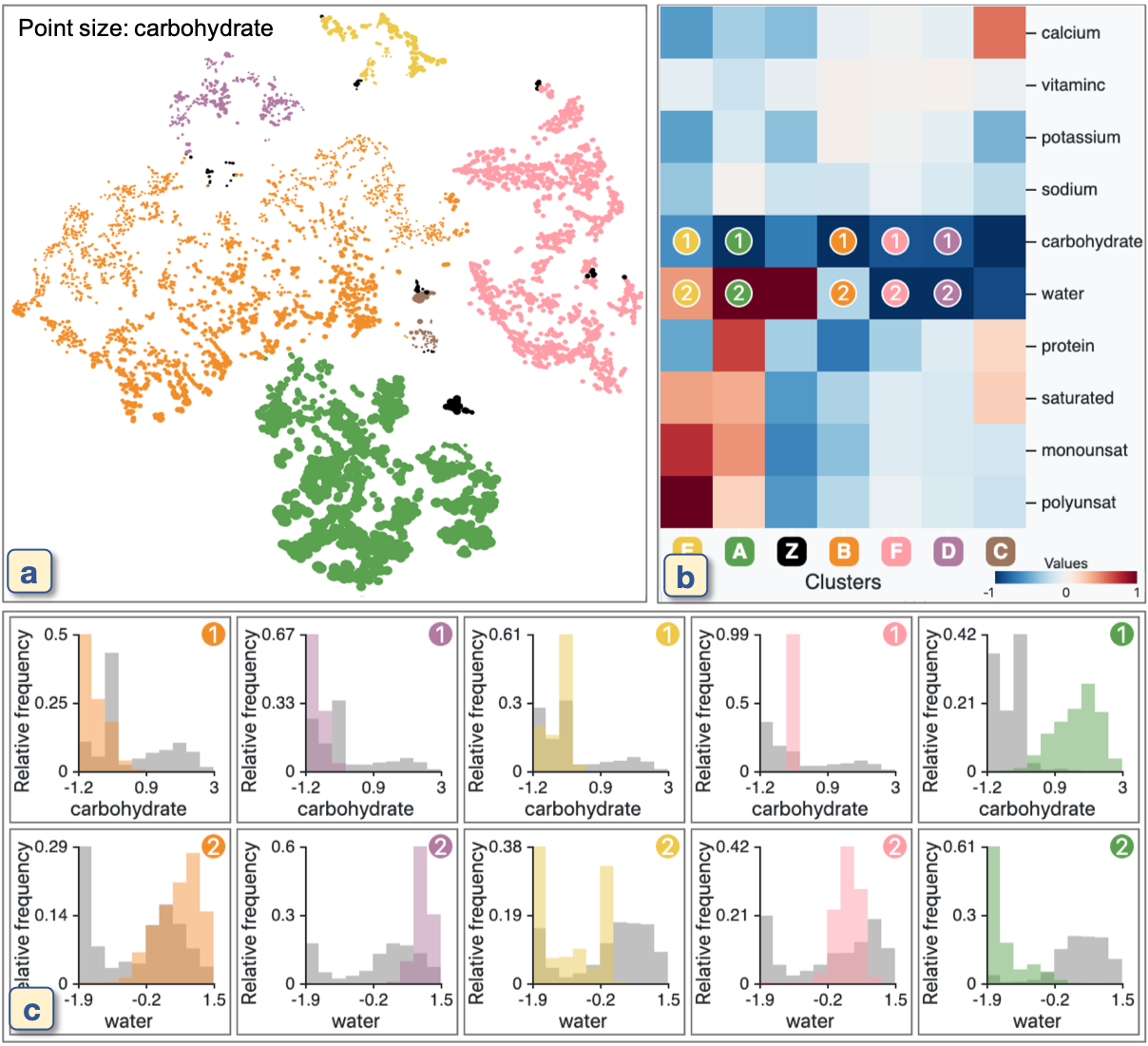}
   	\caption{The result after filtering out the `calories' and `fat' features from the Nutrient dataset. In (a), a point size represents the value of `carbohydrate'. The histograms of the selected cells in\,(b)\,are\,shown\,in\,(c).}
	\label{fig:nutrients_2}
\end{figure}

\subsection{Food and Nutrient}
\label{sec:nutrients}
We analyze the Nutrient dataset in the USDA Food Composition Databases~\cite{usda_food} as an analysis example with a large number of data points.
We use the version available from \cite{nutrient_explorer}. 
This dataset consists of the nutrient content for each food. 
The dataset has 7,637 data points (foods) and 14 features (nutrients). 

This dataset has 12,507 missing values and this is 11.7\% of all the values.
Since this high percentage of missing values could affect an analysis result~\cite{acuna2004treatment}, we first preprocess the dataset to reduce this ratio to less than 5\%~\cite{acuna2004treatment}.
We remove features where more than 40\% of the values are missing.
Also, we remove data points where more than 40\% of the feature values are missing.
Afterward, 7,499 data points and 12 features remain and there are 4,447 missing values (4.9\% of all the values). 
We replace the missing values with the mean of each corresponding feature. 

The result after using t-SNE, DBSCAN, and our methods is shown in \autoref{fig:nutrients_1}.
As shown in \autoref{fig:nutrients_1}b, we can see that all clusters except for the brown cluster have high FCs in `calories', `fat', or both. 
When comparing the histograms of `calories' and `fat' for each cluster, as shown in \autoref{fig:nutrients_1}c, each cluster, in fact, has different distributions in `calories' and `fat'.
For example, while the yellow cluster tends to have low calories and fat, the orange cluster tends to have\,high\,values\,for\,both.

We have understood the main characteristics of each cluster.
However, the effects of the two specific features  (`calories' and `fat') are too dominant.
As a result, we cannot find any other interesting patterns. 
We preprocess the dataset to filter out these two features and generate a new result with new cluster labels, as shown in \autoref{fig:nutrients_2}.
At this time, from \autoref{fig:nutrients_2}b, we can see that most of the clusters are contrasted by mainly `water', `carbohydrate', or both. 
For example, the purple and orange clusters placed in the upper left of \autoref{fig:nutrients_2}a have fewer carbohydrates and more water when compared with the pink and green clusters, as shown in \autoref{fig:nutrients_2}c.
These two examples show that the FCs are useful to know which features have a dominant effect on cluster forming in the DR result.

\begin{figure}[tb]
    \captionsetup{farskip=0pt}
	\centering
	\hspace*{-4pt}
	\subfloat[DR result]{
     \includegraphics[height=1.025\linewidth]{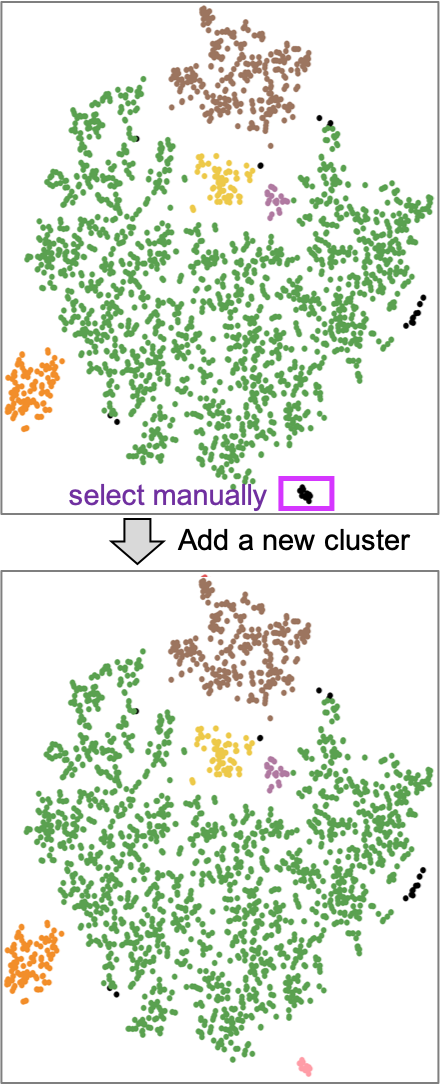}
     \label{fig:crime_dr}
    }
    \hspace*{-4pt}
    \subfloat[Aggregated FCs and histograms]{
     \includegraphics[height=1.025\linewidth]{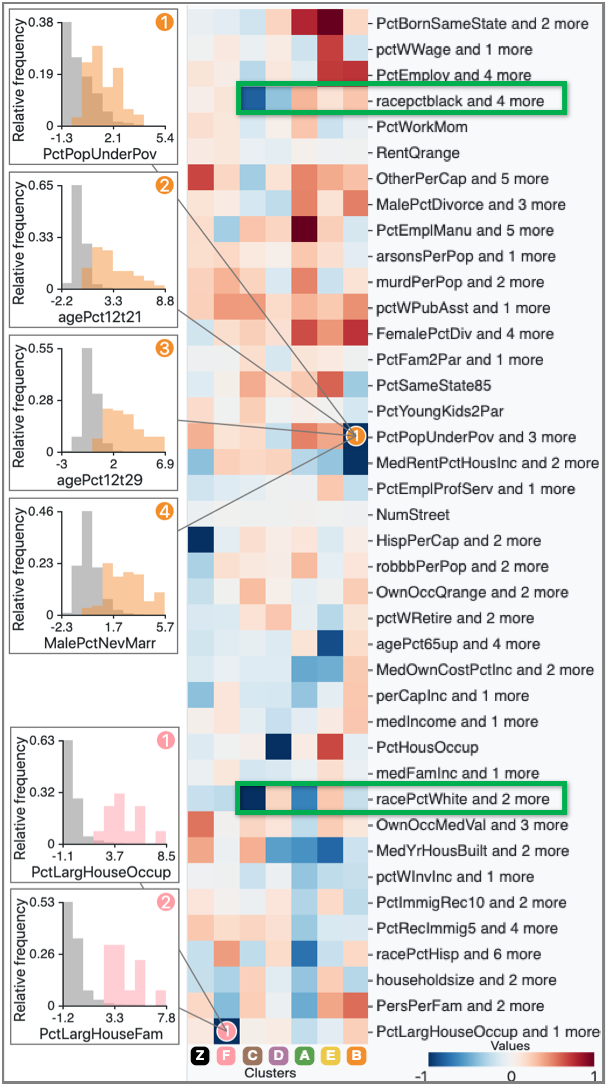}
     \label{fig:crime_fc}
    }
    \vspace{2pt}
   	\caption{The results for the Communities and Crime dataset. The top of (a) shows the result with t-SNE and DBSCAN. In the bottom of (a), the pink cluster which was not identified by DBSCAN is manually added. (b) shows 40 aggregated features from 121 features. Also, some of the histograms of the original features are visualized at the left of (b).}
	\label{fig:crime}
\end{figure}

\subsection{Communities and Crime}
As an example with a large number of features, we analyze the Communities and Crime dataset~\cite{redmond2002data} from \cite{uci_mlr}.
This dataset consists of both socio-economic and crime statistics (e.g., the median family income and the number of murders) for each community.
The dataset contains 2,215 data points (communities) and 143 features (statistics) after excluding identifiers (e.g., county codes).

Because this dataset has many missing values (42,147 values, 13.3\% of all the values), as similar to \autoref{sec:nutrients}, we remove the features where more than 80\% of the values are missing. 
The dataset now has 121 features and only 963 missing values (0.4\% of all the values). 
We replace the missing values with the mean of each corresponding feature. 

\autoref{fig:crime}a (top) shows the result after DR and clustering.
As indicated with the purple rectangle, we manually select an additional cluster as a pink cluster.
Then, we obtain the FCs, as shown in \autoref{fig:crime_fc}.
Because there are many features, the system has aggregated them into 40 features using the aggregation method described in \autoref{sec:aggregation}. 
From \autoref{fig:crime_fc}, we can say that the small clusters (yellow, purple, brown, orange, and pink) are separated from the green cluster due to race, house size, etc.---not due to the criminal statistics. 
For instance, as indicated with the green rectangles, the brown cluster has high FCs in race percentages of African Americans and Caucasians (`racepctblack' and `racePctWhite').
Also,\,the\,pink cluster has high FC in `PctLargHouseOccup' (percentage of all occupied households that are large).

We show the histograms of the features aggregated to the `PctLargHouseOccup and 1 more', as shown in the lower left of \autoref{fig:crime_fc}.
We can see that both `PctLargHouseOccup' and `PctLargHouseFam' (percentage of family households that are large) have similar distribution patterns. 
These patterns can be found because our aggregation method is performed after applying the optimal sign flipping and ordering described in \autoref{sec:features_contributions_view}.
Our aggregation method is able to provide a scalable visualization and help the user analyze many features.
Another example for `PctPopUnderPov and 3 more' of the orange cluster is shown in the upper left of \autoref{fig:crime_fc}.
All `PctPopUnderPov' (percentage of people under the poverty level), `agePct12t21' (percentage of population that is 12--21), `agePct12t29' (percentage of population that is 12--29), and `MalePctNevMarr' (percentage of males who have never married) tend to have a higher value in comparison to that of others. 

\section{Discussion and Limitations}
\textbf{Generality of our method.}
We utilize cPCA~\cite{abid2017contrastive,abid2018exploring} to find features contrasting the target cluster.
We discuss the reason why we use this approach instead of analyzing how the DR method generates clusters. 
If possible, the latter approach would be effective because the cluster formation is a result of the DR method.
However, many of the nonlinear DR methods used for visualization (e.g., t-SNE~\cite{maaten2008visualizing}, LargeVis~\cite{tang2016visualizing}, and UMAP~\cite{mcinnes2018umap}) generate irreversible low-dimensional projection of the original data structure. 
These methods do not have a parametric mapping between the original and projected dimensions; therefore, it is difficult to provide information about how these DR methods affect cluster forming. 
Our methods provide flexibility for analyzing results from any type of DR methods.

We introduce using cPCA to understand the characteristics of the clusters identified in the DR result. 
Our methods can also be used in other situations.
For example, even though using DR before clustering is a common approach~\cite{xu2005survey,wenskovitch2017towards}, our methods can support visual analytics of clusters that are obtained from the clustering methods without going through the DR step. 
This would be helpful to understand clusters' characteristics and to analyze the quality of the clustering methods without any effects derived from DR (e.g., distortion in the projection space). 
Another example is applying our methods to labeled data.
Our methods can identify the essential features to contrast a labeled group from the others.
Therefore, our methods would be useful to understand the characteristics of each group and could help design classifiers based on the gained knowledge.
Our prototype system can support these types of analysis by changing the parts related to steps (a) and (b) in \autoref{fig:workflow}, such as the DR view and clustering algorithms.

\vspace{3pt}
\noindent\textbf{Advantages of using cPCA.}
In \autoref{sec:find_features}, we have already discussed the advantages of using cPCA when compared with using PCA and LDA. 
It is also possible to compute the discrepancy score $D$ introduced in \autoref{sec:automatic_selection_alpha} for each original feature without using ccPCA and then use the score as the feature contribution. 
However, this approach has a similar problem with LDA because the obtained score only shows the separation and does not take into account the variety (i.e., variance) for each feature.

Another potential option is using the two-group differential statistics methods~\cite{marusteri2010comparing}, such as two-sample t-test, Wilcoxon signed-rank test, and Mann-Whitney U test, to find features that have differences between the target cluster and the others. 
Unlike LDA, PCA, or cPCA, these methods cannot produce a quantitative measure for analyzing the FCs to the contrast of the cluster.
More importantly, these statistical methods are designed to test whether there is a difference in a certain statistic (typically mean) between two clusters.
Therefore, these methods are not suitable for performing exploratory analysis on clusters when we do not know their characteristics beforehand.

\vspace{3pt}
\noindent\textbf{Limitations.}
Since we use cPCA, we will need to address its limitations in terms of time and space complexity for a large scale problem.
Similar to the classical PCA, cPCA computes the covariance matrices and then performs EVD. 
For a fixed $\alpha$, it has the same time and space complexity with PCA, which are $\smash{O(d^2n + d^3)}$ and $\smash{O(d^2)}$, respectively, where $n$ is the number of data points and $d$ is the number of features.
Thus, cPCA can achieve fast computation for a dataset which has a large $n$, but not for a dataset with a large $d$ (we include the experimental results in the Supplementary Materials~\cite{supp}).
For PCA, incremental algorithms~\cite{oja1985stochastic,weng2003candid,ross2008incremental} have been developed to solve this issue.
For example, the algorithm in \cite{ross2008incremental} has the time and space complexity of $\smash{O(dm^2)}$ and $\smash{O(d(k + m))}$, respectively, where $m$ is the number of data points used in each batch, and $k$ is the number of principal components. 
We thus plan to develop an incremental version of cPCA next. 

\section{Conclusions}

Dimensionality reduction is widely used to analyze high dimensional data for pattern discovery and real-world problem-solving.
Our work makes a tangible contribution to interpreting and understanding DR results by introducing a visual analytics method that capitalizes on contrastive learning.
Using a scalable visualization, the method directs the user to the essential features within the data.
Our work, thus, further enhances the usability of DR methods. 

\acknowledgments{The authors wish to thank Suyun Bae (suybae@ucdavis.edu) of VIDI Labs at the University of California, Davis, for her assistance in improving the clarity of the paper content. This research is sponsored in part by the U.S.~National Science Foundation through grants IIS-1528203 and IIS-1741536.}

\bibliographystyle{abbrv-doi}
\bibliography{00_main}

\end{document}